\documentclass[runningheads]{llncs}

\usepackage{eccv}

\usepackage{eccvabbrv}
\usepackage{wrapfig}
\usepackage{graphicx}
\usepackage{booktabs}

\usepackage[accsupp]{axessibility}  
\usepackage{multirow}

\usepackage{hyperref}

\usepackage{orcidlink}
\DeclareMathOperator*{\argmin}{arg\,min}

\begin{document}

\title{HyperSpaceX: Radial and Angular Exploration of HyperSpherical Dimensions} 

\titlerunning{HyperSpaceX}

\author{Chiranjeev Chiranjeev\orcidlink{0000-0003-3026-1255} \and
Muskan Dosi\orcidlink{0000-0001-7451-3317} \and
Kartik Thakral\orcidlink{0000-0002-2528-9950} 
Mayank Vatsa\orcidlink{0000-0001-5952-2274} \and
Richa Singh\orcidlink{0000-0003-4060-4573}}

\authorrunning{Chiranjeev et al.}

\institute{Indian Institute of Technology Jodhpur, India
\email{\{chiranjeev.1,dosi.1,thakral.1,mvatsa,richa\}@iitj.ac.in} \\
\textcolor{magenta}{\url{https://github.com/IAB-IITJ/HyperSpaceX}}}

\maketitle

\begin{center}
    \captionsetup{type=figure}
    \includegraphics[width=0.79\textwidth]{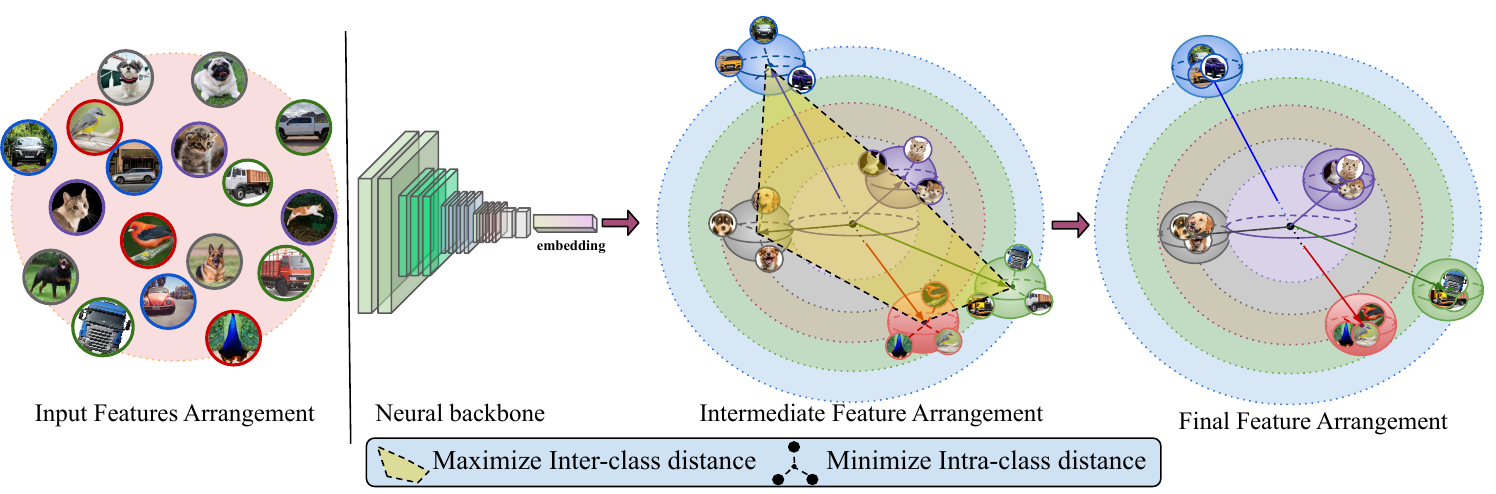}
    \captionof{figure}{Visual concept of the proposed HyperSpaceX framework: It utilizes a novel radial-angular latent space on hyperspherical manifolds to differentiate features effectively. Initially, class features are indistinguishable due to overlap. The proposed DistArc loss exhibits two feature arrangement learning: high inter-class variation employing multi-radial angular arrangement and minimal intra-class distance, leading to highly separable-discriminable class features.} 
\end{center}

\begin{abstract}
 Traditional deep learning models rely on methods such as softmax cross-entropy and ArcFace loss for tasks like classification and face recognition. These methods mainly explore angular features in a hyperspherical space, often resulting in entangled inter-class features due to dense angular data across many classes. In this paper, a new field of feature exploration is proposed known as \textit{HyperSpaceX} which enhances class discrimination by exploring both angular and radial dimensions in multi-hyperspherical spaces, facilitated by a novel \textit{DistArc} loss. The proposed DistArc loss encompasses three feature arrangement components: two angular and one radial, enforcing intra-class binding and inter-class separation in multi-radial arrangement, improving feature discriminability. Evaluation of \textit{HyperSpaceX} framework for the novel representation utilizes a proposed predictive measure that accounts for both angular and radial elements, providing a more comprehensive assessment of model accuracy beyond standard metrics. Experiments across seven object classification and six face recognition datasets demonstrate state-of-the-art \textit{(SoTA)} results obtained from \textit{HyperSpaceX}, achieving up to a 20\% performance improvement on large-scale object datasets in lower dimensions and up to 6\% gain in higher dimensions. 
 \keywords{Representation Learning \and Image Classification \and Face Recognition}
\end{abstract}

\section{Introduction}
\label{sec:intro}

The advancement in image classification and face recognition has greatly benefited from the introduction of innovative loss functions, designed to better differentiate between class features. These functions are crafted to learn unique representations for each class by employing proxies or weight vectors to increase the distinction between classes. Traditional image classification techniques \cite{lecun1989backpropagation}, \cite{krizhevsky2012imagenet},\cite{sermanet2013overfeat}, \cite{zeiler2014visualizing} primarily rely on cross-entropy loss \cite{conniffe1987expected}, \cite{zhang2018generalized} to create intricate class boundaries and improve model generalization. However, a common limitation of these approaches is their tendency to neglect the reduction of within-class feature distances while failing to adequately separate features between different classes. This results in the blending of feature points from dissimilar classes, thus leading to reduced accuracies in object classification and face verification tasks. 

To address this gap, several modifications to softmax-based loss functions have been introduced, utilizing proxy-based methods to cluster similar class features while concurrently expanding the separation between dissimilar ones within the latent space. For face recognition, existing approaches \cite{deng2019arcface}, \cite{guo2019survey}, \cite{liu2017sphereface}, \cite{sun2014deep}, \cite{wang2018cosface} leverage proxy-based objectives to achieve better feature separability on a hypersphere through angular dimensions. Methods such as SphereFace \cite{liu2017sphereface}, \cite{Wen2022SphereFace2} and ArcFace \cite{deng2019arcface} introduce angular margins to promote large-margin discriminative feature learning, while CosFace \cite{wang2018cosface} incorporates marginal cosine functions to modify softmax loss by equalizing radial variations, introducing a cosine margin for angular decision boundaries. Furthermore, face recognition has benefited from proxy-free training strategies that focus on deep metric learning techniques \cite{oh2016deep}, \cite{schroff2015facenet}, \cite{sohn2016improved}, \cite{wu2017sampling}. These strategies generate embeddings that bring similar faces closer together while keeping dissimilar ones apart, adapting effectively to various conditions without the need for explicit class labels. However, methods like triplet learning \cite{schroff2015facenet} and contrastive loss \cite{wu2017sampling} depend heavily on the availability of extensive image pairs or triplets for effective training and often face difficulties with hard-pair mining.  Other discriminative loss functions like Center loss \cite{centerloss} and Git loss \cite{gitloss} encounter difficulties in updating class center parameters with a large identity count, adding computational cost. Orthogonal Projection Loss (OPL)\cite{opl} and Learnable Subspace Orthogonal Projection \cite{subspaceop} employ orthogonal projection constraints to improve separability between classes, but this approach restricts the number of classes to $2^{d}$ in d-dimensional space.

The limitations of traditional loss functions, including those modified for angular space, have led to challenges in effectively differentiating between features. The issue occurs when features of closely related classes overlap, causing uncertainties in the correct identification or classification of subjects\footnote{It is further analyzed in section \ref{sec:evaluation_metric}}. This overlap of class features makes it difficult to discern between unique identities or classes, thereby impacting the overall performance. To overcome these limitations, we propose the \textit{HyperSpaceX} framework, a novel approach that extends the exploration of feature space beyond the angular to include radial dimensions in multi-hyperspherical latent spaces. It introduces a loss function that emphasizes the arrangement of features to increase inter-class separability and intra-class compactness. Through the effective combination of angular and radial organization of features, the framework aims to minimize the overlap among inter-class data points, presenting a novel discriminative feature representation  approach for image classification and face recognition. The key highlights are:
\begin{enumerate}
     \item Introducing \textit{HyperSpaceX}, a novel discriminative feature representation and arrangement learning that explores both radial and angular dimensions within multi-hyperspherical spaces, enabling effective feature learning.
    
     \item Developing the \textit{DistArc} loss function, aimed at enhancing the discriminative power of deep learning models. It improves feature representation by promoting better separation between different classes and tighter clustering of features within the same class in the latent space.

     \item Presenting a predictive measure to evaluate the model's performance, incorporating both radial and angular dimensions in multi-hyperspherical settings. This measure provides a more comprehensive understanding of the model's capabilities in handling complex feature distributions.
    \item Evaluating the proposed approach using seven object datasets (MNIST, FashionMNIST, CIFAR-10, CIFAR-100, CUB-200, TinyImageNet and ImageNet1K) and six face datasets (LFW, CFP-FP, AgeDB-30, CA-LFW, CP-LFW and D-LORD), showing its effectiveness in improving feature distinction and model accuracy across various data types and complexities, often achieving state-of-the-art results.    
\end{enumerate}

\section{The HyperSpaceX Framework}
\label{sec:proposed}

In traditional hyperspherical angular latent spaces, identities or classes are delineated by distinct angular directions. However, with the increase in the number of classes and the volume of data for each class, overlaps and intersections in these class representations increases, leading to significant classification challenges, as they blur the distinction between different classes. To address the issue of feature overlap and the necessity for more defined class distinctions, we propose the HyperSpaceX framework. This framework utilizes a multi-hyperspherical space, drawing on both radial hyperspheres and angular dimensions to promote a discriminative distribution of class features. This learnable arrangement distribution is facilitated by the novel DistArc loss that significantly enhances class differentiation and separation along with the grouping of same-category features. DistArc strategically arranges features by considering both their angle and distance from the center, across various spherical layers. This method utilizes the diversity of spatial dimensions to establish clear and distinct areas within the spherical model, making it easier to define precise boundaries between classes. Such a multi-dimensional approach generates a more elaborate hyperspherical subspace, facilitating the creation of more efficient decision boundaries.

\subsection{Preliminaries}
Deep learning research has explored a variety of loss functions, among which Cross-entropy loss is most commonly used for classification tasks. Let $x_{i}$ and $\omega$ be the feature embedding and weight or proxy matrix, respectively. Cross-entropy loss is defined as $L_{\text{CE}} = - \; \frac{1}{N} \sum_{i=1}^N \log \frac{e^{\omega^{T}_{y_{i}} x_{i} + b_{y_{i}}}}{\sum_{j=1}^K e^{\omega^{T}_{j} x_{j} + b_{j}}}$.  Here, $\omega_{y_{i}}$ signifies the proxy vector of the $y_{i}$-th class. The term $b$ refers to the bias, $N$ represents the number of samples distributed over $K$ number of classes. This function calculates the expected loss between the predicted and actual class distributions, focusing on optimizing class separations without necessarily bringing similar class features closer or significantly distancing different classes. Such limitations can adversely affect image classification and face recognition performance, particularly under high intra-class variations. There have been attempts to enhance inter-class distinctions through loss functions like Orthogonal Projection Loss (OPL) \cite{opl}, which introduces perpendicular margins between class features. However, these efforts are often overshadowed by the dominance of cross-entropy in optimization processes. Softmax-based losses, including those used in deep face recognition like SphereFace \cite{liu2017sphereface} \cite{Wen2022SphereFace2}, CosFace \cite{wang2018cosface} and ArcFace \cite{deng2019arcface}, primarily rely on cross-entropy based formulation. Among these is ArcFace loss, one of the most popular angular-loss functions, which distinguishes itself by allocating features around class proxies within a hyperspherical space and applying an angular margin $m$ to enhance the cohesion within classes and the distinction between different classes. The mathematical formulation of ArcFace loss is represented as $L_{\text{ArcFace}} = - \; \frac{1}{N} \sum_{i=1}^N  \log \frac{e^{cos(\theta_{y_{i}}+m)}}{e^{cos(\theta_{y_{i}}+m)} + \sum_{j=1, j \neq y_{i}}^{K} e^{cos\theta_{j}}}$.

\noindent In this, $cos\theta$ is calculated using cosine similarity between feature embedding $x$ and proxy $\omega$, that is defined by computing a dot product between unit vectors of $x$ and $\omega$ as represented through $cos(\theta_{y_{i}}) = \hat{x}_{i} \cdot \hat{\omega}_{y_{i}} \,\; \;\, cos(\theta_{j}) = \hat{x}_{i} \cdot \hat{\omega}_{j}$. However, focusing solely on angular dimensions can lead to ambiguity among classes in densely populated identity spaces, highlighting a need for broader exploration beyond angular metrics.

\subsection{Proposed DistArc Loss}
\label{proposed loss}

\begin{figure}[!t]
    \centering
    \includegraphics[scale=0.57]{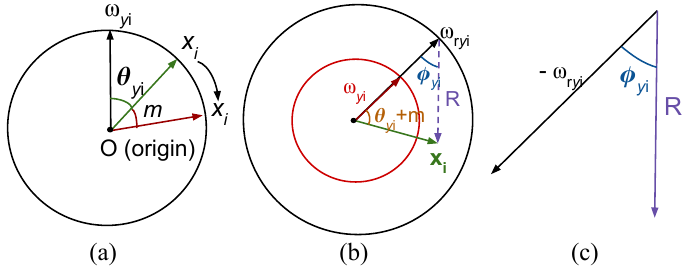}
    \caption{The geometric interpretation of angular and radial formations in multi-hyperspherical dimensions in the training phase through (a) $\theta$ and angular-margin penalty \textit{m}, and (b) angle $\phi$ between scaled proxy vector $\omega_{y_{i}}$ and resultant vector $R$. (c) Shows the vector representation of $R$ and $\omega_{y_{i}}$ in a reverse direction for computing angle $\phi$ using cosine of an angle $\phi$.}
    \label{fig:train_geom}
\end{figure}

The proposed \textit{DistArc} loss navigates through radial and angular dimensions across multiple hyperspheres centered at the origin to improve class discrimination. The \textit{DistArc} loss is formulated as, 
\begin{equation}
L_{\text{DistArc}} = - \frac{1}{N} \sum_{i=1}^N  \log \frac{e^{\cos(\theta_{y_{i}}+m) \; + \; \cos(\phi_{y_{i}}) \; - \; \lambda \delta_{y_{i}}}}{e^{\cos(\theta_{y_{i}}+m)} + \sum_{j=1, j \neq y_{i}}^{K} e^{\cos(\theta_{j}) \; - \; \lambda \delta_{j}}}
\label{eq:distarc_loss}
\end{equation}
where $\lambda$ represents the weighing factor and $\cos\theta$ defines the cosine of angle between embedding $x$ and a proxy vector $\omega$. An angular-margin penalty of $m$ is introduced in $\cos\theta$ to enhance the feature discriminability. \cref{fig:train_geom}a represents the geometric illustration of $\cos(\theta_{y_{i}}+m)$. The DistArc loss minimizes the angle $\theta_{y_{i}}$, thereby increasing the cosine similarity between embeddings $x_{i}$ and their corresponding proxy's $\omega_{y_{i}}$. However, to include the radial dimension also we incorporate the $\cos\phi_{y_{i}}$ which represents the cosine of angle between radii scaled proxy vector $\omega_{r_{y_{i}}}$ and a resultant vector $R_{y_{i}}$ of $\omega_{r_{y_{i}}}$ and ${x_{i}}$. \cref{fig:train_geom}b describes the geometric representation of the resultant vector $R_{y_{i}}$. $R_{y_{i}}$ is modified while learning/training such that it minimizes the angle between $R_{y_{i}}$ and $\omega_{r_{y_{i}}}$ which is defined using $ \label{eq:Wr} \omega_{r_{y_{i}}} =  \hat{\omega}_{y_{i}} \times {r_{y_{i}}}$ and $R_{y_{i}}$ is defined as $R_{y_{i}} = \, - \, \omega_{r_{y_{i}}} + \: x_{i}$. Here, $R_{y_{i}} \in \mathbb{R}^{\textit{d}}$, $\omega_{r_{y_{i}}} \in \mathbb{R}^{\textit{d} \times \textit{1}}$, and $x_{i} \in \mathbb{R}^{\textit{d}}$. The resultant vector helps to maintain the embedding magnitude within the hyperspherical radii of magnitude $||\omega_{r_{y_{i}}}||_{2}$ and further optimizes the embeddings $x_{i}$\textit{'s} to cluster in the angular direction by minimizing the angle $(\phi_{y_{i}})$ between $R_{y_{i}}$ and $\omega_{r_{y_{i}}}$: \pmb{$\cos(\phi_{y_{i}}) = \hat{R}_{y_{i}} \cdot -\hat{\omega}_{r_{y_{i}}}$}. This equation, while calculating the $ \cos\phi_{y_{i}}$, the origin is taken at $\hat{\omega}_{r_{y_{i}}}$ which results $\hat{\omega}_{r_{y_{i}}}$ vector to be in reverse direction leading to $-\hat{\omega}_{r_{y_{i}}}$. We can visualize its vector representation from \cref{fig:train_geom}c representing the angle computation process between the reversely directed scaled proxy vector $\omega_{r_{y_{i}}}$ and a resultant vector $R_{y_{i}}$. The term $\cos\phi_{y_{i}}$ within the proposed loss function $L_{DistArc}$ is designed to enhance the intra-class compactness of features with their respective scaled class proxies, $\omega_{r_{y_{i}}}$ by optimizing both the radial and angular dimensions of different hyperspheres. This term also has a major significance in letting the distribution of each class to stay within particular radii hyperspheres in the feature space. 

Through \textit{DistArc} loss, the $\cos\theta_{y_{i}}$ optimizes to align the embeddings with class proxies $\omega_{y_{i}}$ in the angular space and $\cos\phi_{y_{i}}$ optimizes to increase the compactness among embeddings and their respective scaled proxies $\omega_{r_{y_{i}}}$. It also ensures that the length of embeddings does not extend beyond their radial space $r_{i}$; further, minimizes the angle between normalized resultant vector $(\hat{R}_{y_{i}})$ and normalized scaled proxy $(\hat{\omega}_{r_{y_{i}}})$ to bring closer the embeddings at the point of scaled proxy, $\omega_{r_{y_{i}}}$ in the angular space. The overall geometric interpretation of \textit{DistArc} loss during the training phase is shown in \cref{fig:train_geom}b.

In the proposed DistArc loss, $\delta_{y_{i}}$ act as a pulling force to attract embeddings $x_{i}$ towards $\omega_{r_{y_{i}}}$. This loss component optimizes to shift the distribution of each class towards the scaled proxies $\omega_{r_{y_{i}}}$ in the radial space, leading to an increase in the embedding's magnitude. Leading to the enhancement of intra-class compactness, $\delta_{j}$ in the denominator of the loss function, helps to distant the $x_{i}$\textit{'s} away from other scaled proxies $\omega_{r_{j}}$\textit{'s} excluding $\omega_{r_{y_{i}}}$. The overall combination of $\delta_{y_{i}}$ and  $\delta_{j}$ increases the inter-class distance between dissimilar class scaled proxies and embeddings while decreasing the intra-class distance between similar class scaled proxy and embeddings. The $\delta_{y_{i}}$ term is defined using, \pmb{$\delta_{y_{i}} = || \omega_{r_{y_{i}}} - x_{i} ||^{2}_{2}$} and in case when $j \neq y_{i}$, \pmb{${\delta_{j} = || \omega_{r_{j}} - x_{i} ||^{2}_{2}}$}. A negative (-) sign is used in the $\delta$ component of loss Eq. \ref{eq:distarc_loss} to maximize the logarithmic distribution overall and subsequently minimize the \textit{DistArc} loss. 

\subsection{Analytical Ablation of DistArc Loss and Inductive Bias for HyperSpaceX Framework}
\label{sec:visualization_proposed_loss_components}
In the proposed \textit{DistArc} loss, the first component $\cos(\theta_{y_{i}}+m)$ represents the cosine of the angle $\theta_{y_{i}}$ between $\omega_{y_{i}}$ and $x_{i}$, incorporating an additional penalty $m$ for the angular margin applied to angle $\theta_{y_{i}}$, to strengthen coherence within the class while increasing differentiation between different classes. The second term, $\cos\phi_{y_{i}}$, calculates the angle between the scaled proxy $\omega_{r_{y_{i}}}$ and its corresponding resultant vector $R_{y_{i}}$. This can be expressed as the dot product between the two vectors: $\omega_{r_{y_{i}}}$ in reverse direction and $R_{y_{i}}$. This aims to minimize the angle $\phi_{y_{i}}$, ensuring that feature points remain close to their scaled proxies and do not extend beyond the boundaries of their respective radial hyperspheres. The last two terms, $\delta_{y_{i}}$ and $\delta_{j}$, help to shift the features to their corresponding scaled proxy $\omega_{r_{y_{i}}}$ points over different radii hyperspheres and diverging $x_{i}$ features from different class proxies $\omega_{r_{j}}$\textit{'s}. It ensures inter-class separability in the radial direction across multi-hyperspherical manifolds. The $\delta_{y_{i}}$ term also results in the tighter clustering of class features around their respective $\omega_{r_{y_{i}}}$\textit{'s} leading to high class discrimination. The combined importance of the components $\cos\theta_{y_{i}}$ and $\cos\phi_{y_{i}}$ manifests itself in the minimization of angles $\theta$ and $\phi$, using a more concentrated angular distribution of features aligning to the corresponding class proxies. Further, it prevents features from extending beyond their respective class radii. The loss's additional $\delta$ component contributes to the compact clustering and length amplification of $x_{i}$ vectors, shifting them to specific hypersphere radii\footnote{A detailed visual and theoretical ablation is included in the supplementary material.}.

Additionally, in the context of \textit{HyperSpaceX} framework, the inductive bias effectively shapes the decision boundaries based on both radial and angular factors. We define the decision boundary as,
\begin{equation}
\label{eq:decision_eq}
    \begin{aligned}
    (\cos(\theta_1 + m) + ||x_1||_{2}) - (\cos(\theta_2) + ||x_2||_{2}) &= 0 \text{ for class 1} \\
    (\cos(\theta_1) + ||x_1||_{2}) - (\cos(\theta_2 + m) + ||x_2||_{2}) &= 0 \text{ for class 2}
    \end{aligned}
\end{equation}

This analysis of inductive bias in decision boundaries is built on binary classification and can effectively be extended to multi-class classification tasks. Consequently, the features learned through the radial-angular DistArc loss exhibit more separable decision boundaries and enhanced discriminative capabilities among class features. A more comprehensive derivation of the decision boundary influenced by inductive bias, elucidating its dependency on angles $\phi$, $\theta$, and resultant vector $R$ is present in the supplemental.

\begin{figure}[t!]
    \centering 
    \includegraphics[scale=0.41]{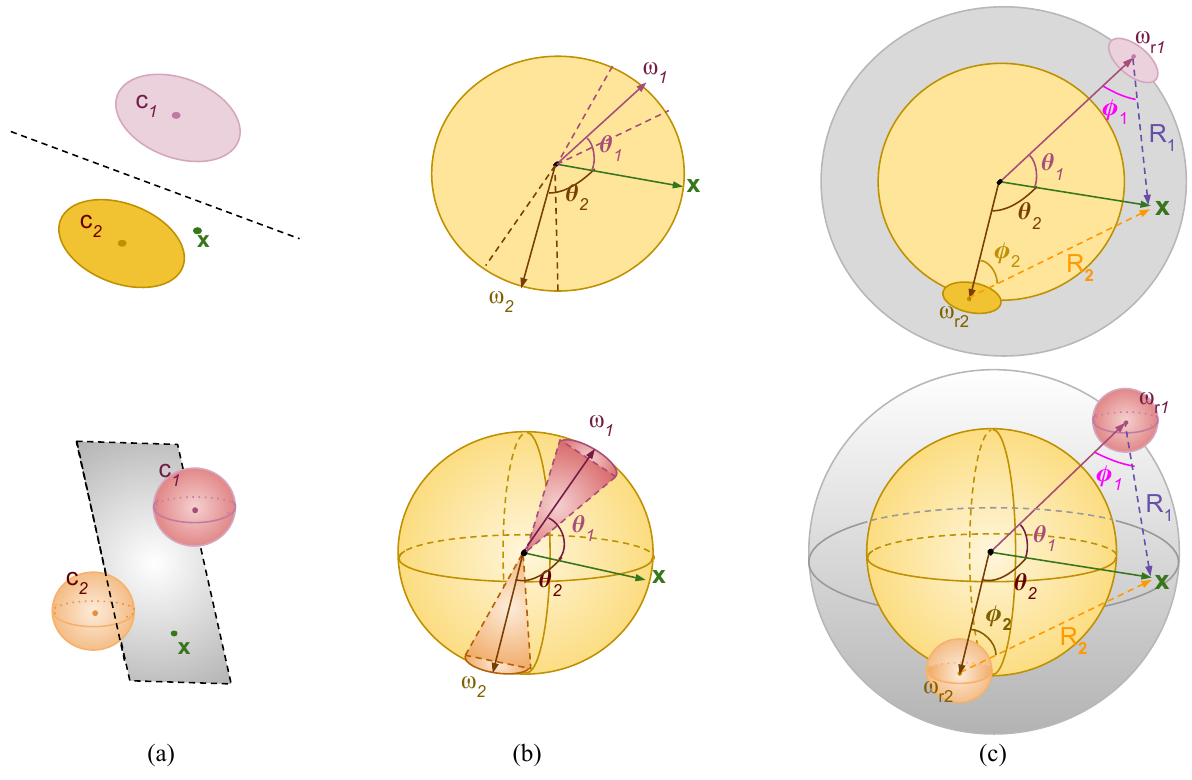}
    \caption{The 2-D (in first row) and 3-D (in second row) latent space visualization of features learnt through (a) metric-based loss functions, (b) Angular-softmax-based loss functions, and (c) the proposed Radial-Angular DistArc loss. Further showing the decision-making process of assigning a test sample $x$ to the most favourable class distribution represented using either class center $c$ or proxy vector $\omega$.}
    \label{fig:test_geom}
\end{figure}

\section{Predictive Measure and Latent Space Visualization}
\label{sec:evaluation_metric}

The most prevailing choice for evaluating and categorizing a sample involves determining the category by calculating the maximum prediction from the model's predicted probability distribution. The proposed framework identifies the most favourable class distribution across multi-hyperspherical manifolds by considering a combination of angular and radial factors. An optimal class prediction is determined through a distinct approach that helps the evaluation metric, taking into account the radial and angular aspects of hyperspheres. The procedure encompasses the computation of resultant vectors between a sample's feature vector and each scaled proxy through mathematical formulation defined as $R_{i} = x - \omega_{r_{i}}$  $(\forall i \in 1,2,3,...)$. The resultant vector with the smallest magnitude determines the most favourable class proximity. 
Cosine law of triangle length computation is employed to determine the length of the resultant vector $R$ based on learned and computed angles $\theta$, $\phi$ and a magnitude of feature vector $x$. Eq. \ref{eq:evaluation_R_magnitude} defines the formulation for $R_{i}$\textit{'s} length computation for every $i^{th}$ category. 
\begin{equation}
    ||R_{i}||_{2} = ||x||_{2} \, \cos\phi_{i} \: + \: ||\omega_{r_{i}}||_{2} \, \cos(\pi-(\theta_{i}+\phi_{i}))
\label{eq:evaluation_R_magnitude}
\end{equation}
We introduce a favourable class determining predictive measure for the \textit{HyperSpaceX} framework with its radial-angular based formulation, \ie,
\begin{equation}
    \hat{y} = \argmin_{R_{m}}\{ R_{m} \in \mathbb{R}^{\textit{K}}: R_{m} \} 
\label{eq:evaluation_metric_eq}
\end{equation}
where $R_{m} \in \mathbb{R}^{\textit{K}}$ refers to a vector comprising magnitude of resultant vectors between a sample's feature vector $x \in \mathbb{R}^{\textit{d}}$ and scaled proxy matrix $\omega_{r} \in \mathbb{R}^{\textit{d} \times \textit{K}}$ representing the radii scaled \textit{K}-class proxies, and $\hat{y}$ defines a predicted class distribution. The predicted class distribution for a specific sample is effectively represented by the class having the smallest resultant vector magnitude in relation to that sample. The derivation of a predictive measure formulation is provided in the supplementary material.

\begin{figure}[tb]
  \centering
  \begin{subfigure}{0.55\linewidth}
    \includegraphics[width=1.1\textwidth]{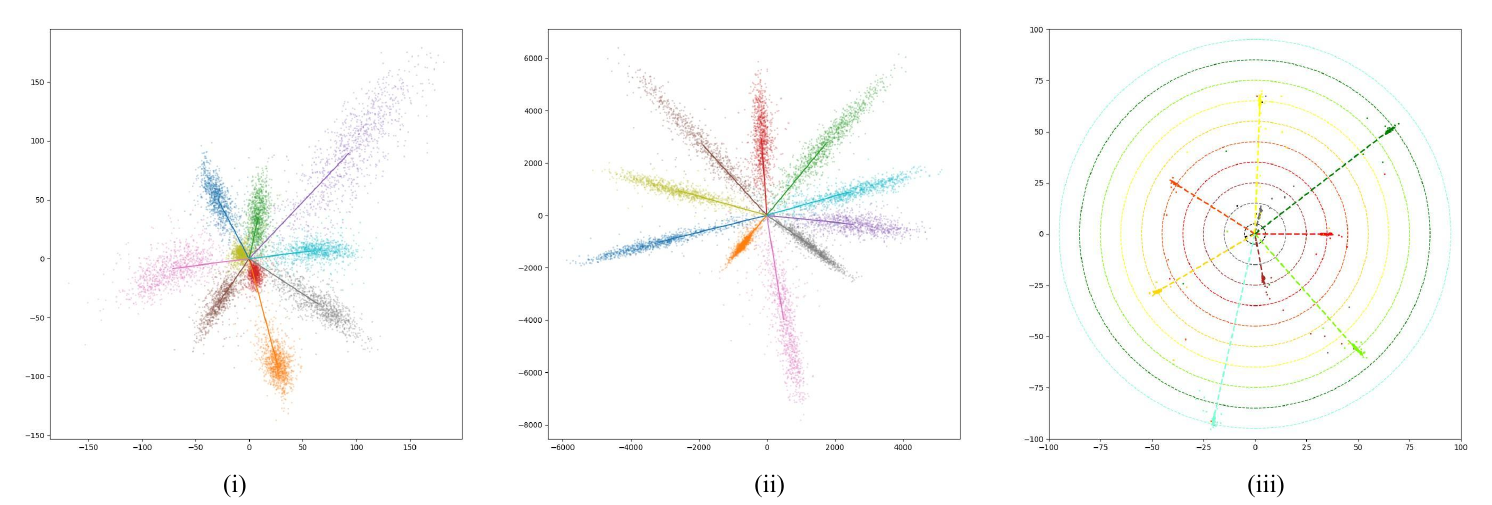} 
    \caption{}
    \label{fig:feature_loss_visualization}
  \end{subfigure}
  \hfill
  \begin{subfigure}{0.42\linewidth}
    \centering
    \includegraphics[scale=0.12]{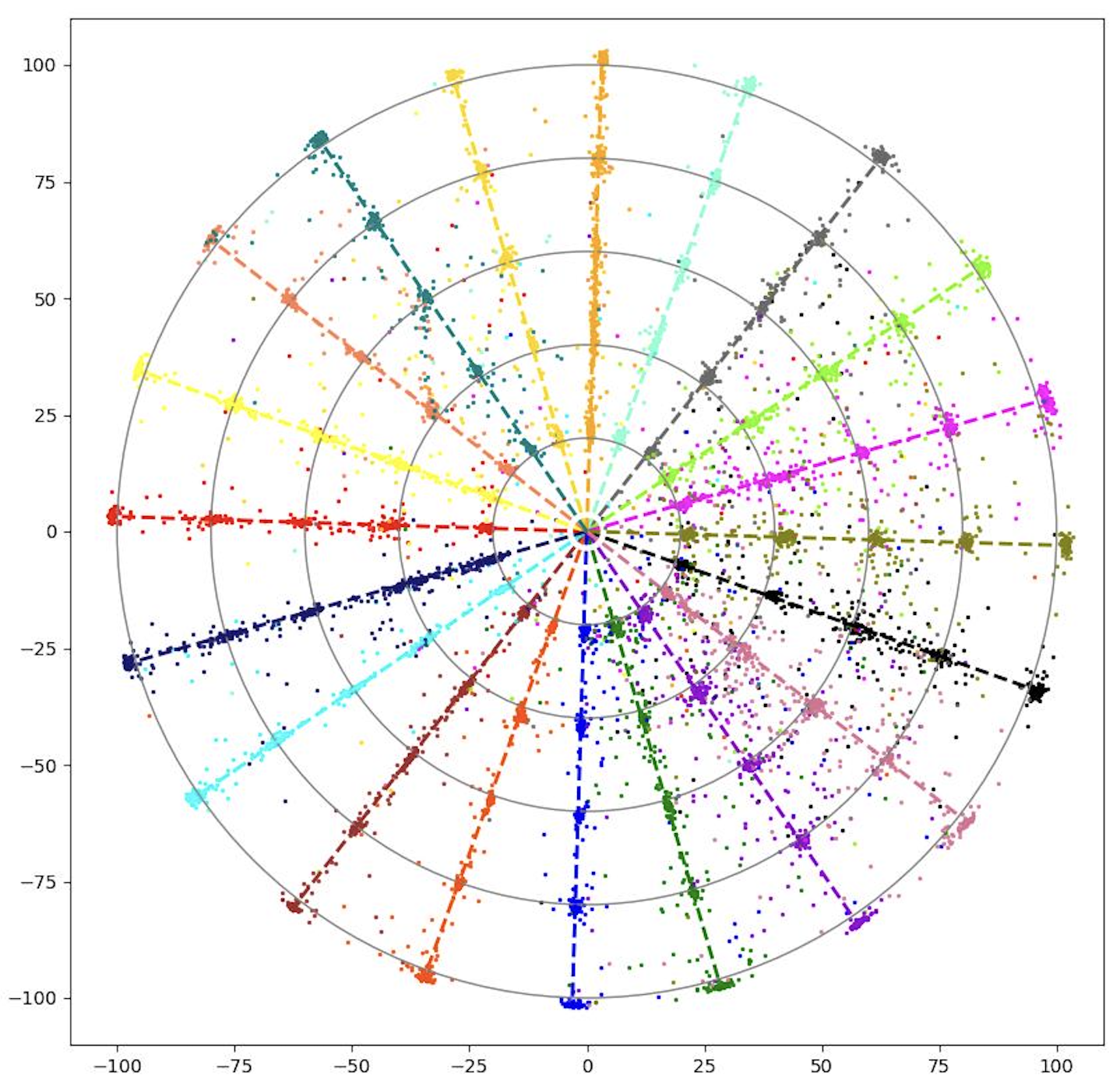}
    \caption{}
    \label{fig:cifar_100_visualization}
  \end{subfigure}
  \caption{(a) Illustrating comparative visual analysis of the organization of MNIST class feature distribution in the latent space. The feature representations are learned using (i) Cross-entropy loss, (ii) ArcFace loss, and (iii) the proposed \textit{DistArc} loss on the MNIST database, where each color represents a unique class. While (b) depicts the subclass organization of the CIFAR-100 dataset on 2D multi-hyperspherical manifolds. The color of each line denotes a distinct superclass, facilitating angular separability. Subclasses within each superclass are distinguished radially, with each subclass represented as blobs extending radially from the superclass center.}
  \label{fig:short}
\end{figure}

During testing, the class of a test sample is determined by its feature representation predicted by the model, influenced by the loss function during training. \cref{fig:test_geom} displays visual representations of 2D and 3D latent space features, elucidating the decision-making process for classifying a test sample. This is achieved by employing a nearest distance class measure, calculated as the Euclidean distance, as shown in \cref{fig:test_geom}a. In \cref{fig:test_geom}b, the class prediction process is shown, based on angular distance within a unit hypersphere. Lastly, \cref{fig:test_geom}c describes the probable class selection process by finding the vector with the least magnitude across multi-hyperspherical manifolds. As shown in \cref{fig:feature_loss_visualization}\textcolor{red}{.i}, the cross-entropy loss facilitates class feature separation, leaving intermixed features that affect prediction confidence due to lack of discriminability. Conversely, angular-softmax losses, introducing angular-margin penalties to enhance cosine similarity among embeddings, aiming separation and discriminability still causes inter-mixed feature representations (depicted via \cref{fig:feature_loss_visualization}\textcolor{red}{.ii}). \textit{HyperSpaceX} framework with \textit{DistArc} loss, presents a solution by distributing features across various angles on multiple hyperspheres, achieving better separation and discriminability by integrating both angular and radial dimensions. This approach results in a more distinct feature arrangement between classes, as demonstrated in \cref{fig:feature_loss_visualization}\textcolor{red}{.iii}), indicating the effectiveness of \textit{DistArc} loss in creating a more discriminable and separable feature space.

\section{Experimental Setup}
The experimental approach is designed to analyze the distribution of features in the latent space across various image classification and face recognition tasks. We have trained and tested the proposed \textit{HyperSpaceX} framework on well-known benchmark datasets: (a) seven image classification datasets including MNIST \cite{deng2012mnist}, FashionMNIST \cite{xiao2017fashion}, CIFAR-10 \cite{krizhevsky2009learning}, CIFAR-100 \cite{krizhevsky2009learning}, CUB-200 \cite{WahCUB_200_2011}, TinyImageNet \cite{le2015tiny} and ImageNet-1K \cite{deng2009imagenet}, and (b) six face recognition datasets including LFW \cite{huang2008labeled}, CFP-FP \cite{sengupta2016frontal}, AgeDB-30 \cite{moschoglou2017agedb}, CA-LFW \cite{calfw}, CP-LFW \cite{cplfw}, and CASIA-WebFace \cite{yi2014learning} is used for initial model training. Model is also trained on $1500$ subjects of D-LORD \cite{manchanda2023d} dataset and tested on $600$ different test subjects. 

\noindent\textbf{Implementation Details:}
\label{sec:implementation}
The HyperSpaceX framework undergoes training using various deep neural network backbones, including the iResNet50 architecture \cite{duta2021improved}, \cite{he2016deep}, the ViT and RN101 backbones \cite{dosovitskiy2020image} from the CLIP foundation model \cite{radford2021learning}. The models are trained using DistArc loss and several variants of softmax-based loss functions, including Cross-entropy, CosFace and ArcFace loss, and also with Orthogonal projection loss. For \textbf{image classification}, the model undergoes both training and testing in a classification setting. We train models for small-class simple datasets like MNIST and FashionMNIST with an SGD optimizer at a learning rate of $1e-2$ and a weight decay of $5e-4$. We further set the value of $m = 0.4$ and $\lambda = 0.003$ in the DistArc loss components. For complex datasets like CIFAR-10, CIFAR-100, CUB-200 and TinyImageNet, the value of $\lambda$ is set to $0.005$. Other hyperparameters remain the same. For \textbf{face recognition}, the training is conducted by employing a classification setup, while during testing, face verification is performed using image pairs with class-set disjoint from training. 
In the face recognition task, the training is performed on the CASIA-WebFace dataset in the classification setting and testing is performed in the verification setup on LFW, CFP-FP, AgeDB-30, CA-LFW and CP-LFW datasets. The SGD optimizer with a learning rate of $1e-3$ and weight decay of $5e-4$ is utilized for both the classifier and backbone network. However, in this task $\lambda$ is varied based on the number of epochs. We initialize by setting $\lambda$ to $0.001$ and increasing it by an addition of $0.001$ after every tenth epoch and stop updating after reaching the value of $0.005$. 

\section{Results and Analysis}
\label{results_and_analysis}

\textbf{Image Classification Results:} We first present the results pertaining to how class features are organized across a range of embedding sizes, from low ($2$D) to high ($512$D) and very high ($2048$D) dimensions, distributed across radial-hyperspherical manifolds. Table \ref{tab:results_image_512_2} showcases the performance on multi-class datasets like CIFAR-100 \cite{krizhevsky2009learning} and TinyImageNet \cite{le2015tiny}, achieved by employing different backbone architectures trained with a variety of loss functions. This highlights the effectiveness of our HyperSpaceX framework, which utilizes DistArc loss for training. This method enhances class distinctiveness, typically outperforming models trained with conventional loss functions such as Cross-entropy, ArcFace, CosFace, and Orthogonal Projection Loss (OPL). These results also showcase that the DistArc loss leads to features with enhanced separability and discrimination due to its effective exploration of both radial and angular dimensions within each hyperspherical latent space. This approach results in a significant performance increase: in TinyImageNet classification tasks, we observe an increase of over 22\% in $2$D, around 1.44\% in $512$D, and 4.93\% in $2048$D spaces, when benchmarked against cross-entropy loss and various softmax-angular margin-based loss functions. A similar trend of substantial improvement is observed for the CIFAR-100 dataset, showing improvements of up to 19.57\% in $2$D, 6.19\% in $512$D, and $2.61$\% in $2048$D spaces.

\cref{fig:cifar_100_visualization} illustrates the subclass and class-based feature distribution for CIFAR-100 categories, arranging each subclass angularly within its superclass group across different radial spaces. This setup enhances radial separability for subclasses and angular separability for superclasses, leading to clear distinctions between them. Furthermore, subclasses are compactly positioned according to their radial and angular coordinates, showcasing precise class feature discrimination. This organized feature distribution demonstrates the capability for a more defined feature arrangement, optimizing representation in lower-dimensional spaces and simplifying model complexity. Consequently, \cref{fig:cifar_100_visualization} uses color differentiation to distinguish between unrelated classes, while same-colored points on different radii indicate subclasses belonging to the same superclass, thereby facilitating a methodical learning of class arrangements in multi-radial spaces for refined differentiation between classes and their subclasses.
 
\begin{table*}[t!]\renewcommand{\arraystretch}{1.2}
\centering
\captionof{table}{\label{tab:results_image_512_2} Performance comparison (accuracy in \%) on image classification task between the proposed DistArc and other loss functions utilized for training models with multiple backbones. The best and second best performances are \textbf{bolded} and \underline{underlined}.}
\resizebox{0.95\textwidth}{!}{
\begin{tabular}{|cl|ccccc|ccccc|}
\hline
\multicolumn{2}{|c|}{\textbf{}}                                                                               & \multicolumn{5}{c|}{\textbf{CIFAR-100}}                                                                                                   & \multicolumn{5}{c|}{\textbf{Tiny-Imagenet}}                                                                                               \\ \hline
\multicolumn{1}{|l|}{\textbf{\begin{tabular}[c]{@{}c@{}}Embedding \\ Size\end{tabular}}} & \textbf{Backbone}  & \textbf{\begin{tabular}[c]{@{}c@{}}Cross\\  Entropy\end{tabular}} & \textbf{ArcFace} & \textbf{CosFace} & \textbf{OPL} & \textbf{DistArc} & \textbf{\begin{tabular}[c]{@{}c@{}}Cross\\  Entropy\end{tabular}} & \textbf{ArcFace} & \textbf{CosFace} & \textbf{OPL} & \textbf{DistArc} \\ \hline
\multicolumn{1}{|c|}{\multirow{4}{*}{\textbf{2}}}                                        & \textbf{iResNet50} \cite{duta2021improved} & 47.12                                                             & {\underline{51.48}}      & 46.92            & 45.20        & \textbf{62.06}   & 14.03                                                             & 16.03            & 13.78            & {\underline{17.73}}  & \textbf{20.89}   \\
\multicolumn{1}{|c|}{}                                                                   & \textbf{RN101} \cite{radford2021learning}     & 37.84                                                             & 36.02            & 35.12            & {\underline{38.85}}  & \textbf{53.95}   & 17.96                                                             & 16.22            & 18.10            & {\underline{ 20.55}}  & \textbf{32.97}   \\
\multicolumn{1}{|c|}{}                                                                   & \textbf{ViT-B} \cite{radford2021learning}    & {\underline{ 58.36}}                                                       & 56.27            & 53.96            & 57.04        & \textbf{77.93}   & 37.58                                                             & {\underline{ 42.94}}      & 36.25            & 36.49        & \textbf{54.11}   \\
\multicolumn{1}{|c|}{}                                                                   & \textbf{ViT-L} \cite{radford2021learning}    & {\underline{67.89}}                                                       & 66.44            & 63.81            & 66.27        & \textbf{79.56}   & 45.93                                                             & 44.06            & {\underline{49.34}}      & 46.20        & \textbf{68.21}   \\ \hline
\multicolumn{1}{|c|}{\multirow{4}{*}{\textbf{512}}}                                      & \textbf{iResNet50} \cite{duta2021improved} & \textbf{83.05}                                                    & 80.22            & 80.01            & 81.28        & {\underline{ 81.62}}      & 61.83                                                             & 60.02            & 58.39            & {\underline{63.06}}  & \textbf{63.27}   \\
\multicolumn{1}{|c|}{}                                                                   & \textbf{RN101} \cite{radford2021learning}    & 69.39                                                             & {\underline{70.84}}      & 68.37            & 69.11        & \textbf{75.58}   & {\underline{ 53.04}}                                                       & 53.01            & 51.77            & 52.85        & \textbf{54.40}   \\
\multicolumn{1}{|c|}{}                                                                   & \textbf{ViT-B} \cite{radford2021learning}    & {\underline{81.99}}                                                       & 81.58            & 79.46            & 80.92        & \textbf{82.62}   & \textbf{78.38}                                                    & 74.22            & 72.61            & 74.42        & {\underline{76.03}}      \\
\multicolumn{1}{|c|}{}                                                                   & \textbf{ViT-L} \cite{radford2021learning}    & {\underline{88.75}}                                                       & 87.27            & 85.40            & 84.83        & \textbf{89.48}   & {\underline{86.49}}                                                       & 85.55            & 81.84            & 85.94        & \textbf{86.79}   \\ \hline
\multicolumn{1}{|c|}{\multirow{4}{*}{\textbf{2048}}}                                     & \textbf{iResNet50} \cite{duta2021improved} & 85.22                                                             & 84.29            & {\underline{86.82}}      & 86.14        & \textbf{89.10}   & 61.39                                                             & {\underline{63.92}}      & 62.40            & 62.71        & \textbf{64.20}   \\
\multicolumn{1}{|c|}{}                                                                   & \textbf{RN101} \cite{radford2021learning}    & {\underline{79.40}}                                                       & 78.15            & 77.06            & 78.22        & \textbf{80.37}   & {\underline{54.13}}                                                       & 53.28            & 53.87            & 54.10        & \textbf{59.06}   \\
\multicolumn{1}{|c|}{}                                                                   & \textbf{ViT-B} \cite{radford2021learning}    & {\underline{81.04}}                                                       & 79.33            & 79.29            & 80.72        & \textbf{83.00}   & {\underline{79.54}}                                                       & 78.37            & 79.26            & 77.00        & \textbf{80.37}   \\
\multicolumn{1}{|c|}{}                                                                   & \textbf{ViT-L} \cite{radford2021learning}    & \underline{87.59}                                                             & 85.93            & 84.74            & {90.20}  & \textbf{90.20}   & 85.03                                                             & 81.39            & 82.48            & {\underline{86.04}}  & \textbf{86.94}   \\ \hline
\end{tabular}}
\end{table*}

The results for the CUB-200 dataset \cite{WahCUB_200_2011}, a large multi-class image classification challenge focusing on bird categorization, are shown in Table \ref{tab:cub_results}. Classifying images in the CUB-200 dataset is particularly demanding due to the subtle, unique features each bird species has, despite the commonality of features like beaks and wings. This demands an emphasis on fine-grained, distinct characteristics of different bird body parts when developing bird image classification models. In such challenging scenario, as shown in Table \ref{tab:cub_results}, the \textit{HyperSpaceX} framework outperforms existing methods. Models trained with DistArc loss achieved a significant performance boost of approximately 20.34\% using a ViT-L backbone with a $2$D last embedding layer. Additionally, DistArc loss trained models excelled in higher-dimensional latent spaces. This success is attributed to our approach's adeptness in discriminately mapping various bird class features across distinctively separated radii and angles in a multi-hyperspherical space, while simultaneously compacting features within the same bird class.

\begin{table*}[t!]\renewcommand{\arraystretch}{1.1}
\centering
\captionof{table}{\label{tab:cub_results} Comparative performance (accuracy in \%) analysis on CUB-200 dataset between proposed DistArc and other loss functions used for training models with multiple backbones. The best and second best performances are \textbf{bolded} and \underline{underlined}.}

\resizebox{0.9\textwidth}{!}{
\begin{tabular}{|c|l|ccccc|}
\hline
\multicolumn{1}{|l|}{\textbf{\begin{tabular}[c]{@{}c@{}}Embedding Size\end{tabular}}} & \textbf{Backbone}  & \textbf{\begin{tabular}[c]{@{}c@{}}Cross Entropy\end{tabular}} & \textbf{ArcFace} & \textbf{CosFace} & \textbf{\begin{tabular}[c]{@{}c@{}}OPL\end{tabular}} & \textbf{DistArc} \\ \hline
\multirow{4}{*}{\textbf{2}}                                                              & \textbf{iResNet50} \cite{duta2021improved} & 7.50                   & 6.19             & 7.07             & \underline{11.59}                                                                            & \textbf{18.09}            \\
                                                                                         & \textbf{RN101} \cite{radford2021learning}    & 11.44                  & 10.47            & 11.42            & \underline{12.84}                                                                            & \textbf{31.50}            \\
                                                                                         & \textbf{ViT-B} \cite{radford2021learning}     & 18.70                  & \underline{20.63}            & 19.04            & 17.06                                                                            & \textbf{25.93}            \\
                                                                                         & \textbf{ViT-L} \cite{radford2021learning}     & 25.48                  & 22.85            & 24.19            & \underline{27.77}                                                                            & \textbf{48.11}            \\ \hline
\multirow{4}{*}{\textbf{512}}                                                            & \textbf{iResNet50} \cite{duta2021improved} & 61.29                  & \underline{63.47}            & 60.29            & 62.83                                                                            & \textbf{70.48}            \\
                                                                                         & \textbf{RN101} \cite{radford2021learning}     & 51.40                  & 50.85            & 50.11            & \underline{54.26}                                                                            & \textbf{58.44}            \\
                                                                                         & \textbf{ViT-B} \cite{radford2021learning}     & 67.93                  & 68.70            & 68.18            & \underline{70.36}                                                                            & \textbf{76.06}            \\
                                                                                         & \textbf{ViT-L} \cite{radford2021learning}    & \underline{80.79}                  & 78.27            & 79.66            & 80.65                                                                            & \textbf{83.28}            \\ \hline
\end{tabular}}
\end{table*}

\noindent \textbf{Reduced Latent Space Representation:} Table \ref{tab:results_image_512_2} and Table \ref{tab:cub_results} represent the performance on several backbones having the last embedding layer of size $2$. It signifies that on complex datasets such as CIFAR-100, CUB-200 and TinyImageNet, the proposed radial-angular feature learning approach in a 2-D latent space achieves a performance boost of 16\% to 20\% over traditional cross-entropy loss methods even with large number of classes. Experimental results on the CUB-200 dataset also depict the high separability among various low-variance classes. Quantitatively, we can deduce that while dimensionality reduction for large complex datasets from a higher dimension of $512$ and $2048$ to the lower dimension of $2$ embedding space, cross-entropy loss observes a $40 - 42\%$ performance gap while DistArc loss successfully reduced this disparity to $20 - 24\%$.

\begin{table*}[t!]\renewcommand{\arraystretch}{1.1}
\centering
\caption{\label{tab:result_both_metrics} Evaluating model performance (accuracy in\%) using conventional and proposed predictive measures in the HyperSpaceX framework when trained using DistArc loss on backbones with 512-D embedding size. The best performances are \textbf{bolded}.}
\resizebox{0.95\textwidth}{!}{
\begin{tabular}{|l|cc|cc|cc|}
\hline
\multirow{2}{*}{\textbf{Backbone}} & \multicolumn{2}{c|}{\textbf{CIFAR-100}} & \multicolumn{2}{c|}{\textbf{CUB-200}} & \multicolumn{2}{c|}{\textbf{TinyImageNet}} \\ \cline{2-7} 
 &
  \textbf{\begin{tabular}[c]{@{}c@{}}Classification \\ layer\end{tabular}} &
  \textbf{\begin{tabular}[c]{@{}c@{}}Radial-Angular\\ (proposed)\end{tabular}} &
  \textbf{\begin{tabular}[c]{@{}c@{}}Classification \\ layer\end{tabular}} &
  \textbf{\begin{tabular}[c]{@{}c@{}}Radial-Angular\\ (proposed)\end{tabular}} &
  \textbf{\begin{tabular}[c]{@{}c@{}}Classification \\ layer\end{tabular}} &
  \textbf{\begin{tabular}[c]{@{}c@{}}Radial-Angular\\ (proposed)\end{tabular}} \\ \hline
\textbf{iResNet50} \cite{duta2021improved}                & 80.29          & \textbf{81.62}         & 69.56                & \textbf{70.48}               & 61.93                & \textbf{63.27}      \\
\textbf{RN101} \cite{radford2021learning}                     & 71.05          & \textbf{75.58}         & 58.19               & \textbf{58.44}               & 54.33                & \textbf{54.43}      \\
\textbf{ViT-B} \cite{radford2021learning}                    & 78.99          & \textbf{82.62}         & 74.91              & \textbf{76.06}             & \textbf{76.10}       & 76.03               \\
\textbf{ViT-L} \cite{radford2021learning}                    & 88.20          & \textbf{89.48}         & 82.85           & \textbf{83.28}               & 82.51                & \textbf{86.79}      \\ \hline
\end{tabular}}
\end{table*}

\noindent \textbf{Predictive Measures Analysis:} In the Image classification task, the class predictions are generally made utilizing the model's classification layer, adopted for models trained with various existing loss functions except for DistArc loss, for results presented in Table \ref{tab:results_image_512_2} and Table \ref{tab:cub_results}.The proposed DistArc loss, the predictions are computed using the newly introduced radial-angular based predictive measure defined in section \ref{sec:evaluation_metric}. The comparison of utilizing traditional and proposed predictive measures in the HyperSpaceX framework is shown via Table \ref{tab:result_both_metrics}. From the outperforming results shown, it can be analyzed that the proposed radial-angular-based predictive measure is effective over the conventional approach of predictions from the final classification layer for models trained using DistArc loss due to different radii of classes.

\begin{figure}[!t]
    \centering
    \includegraphics[scale=0.33]{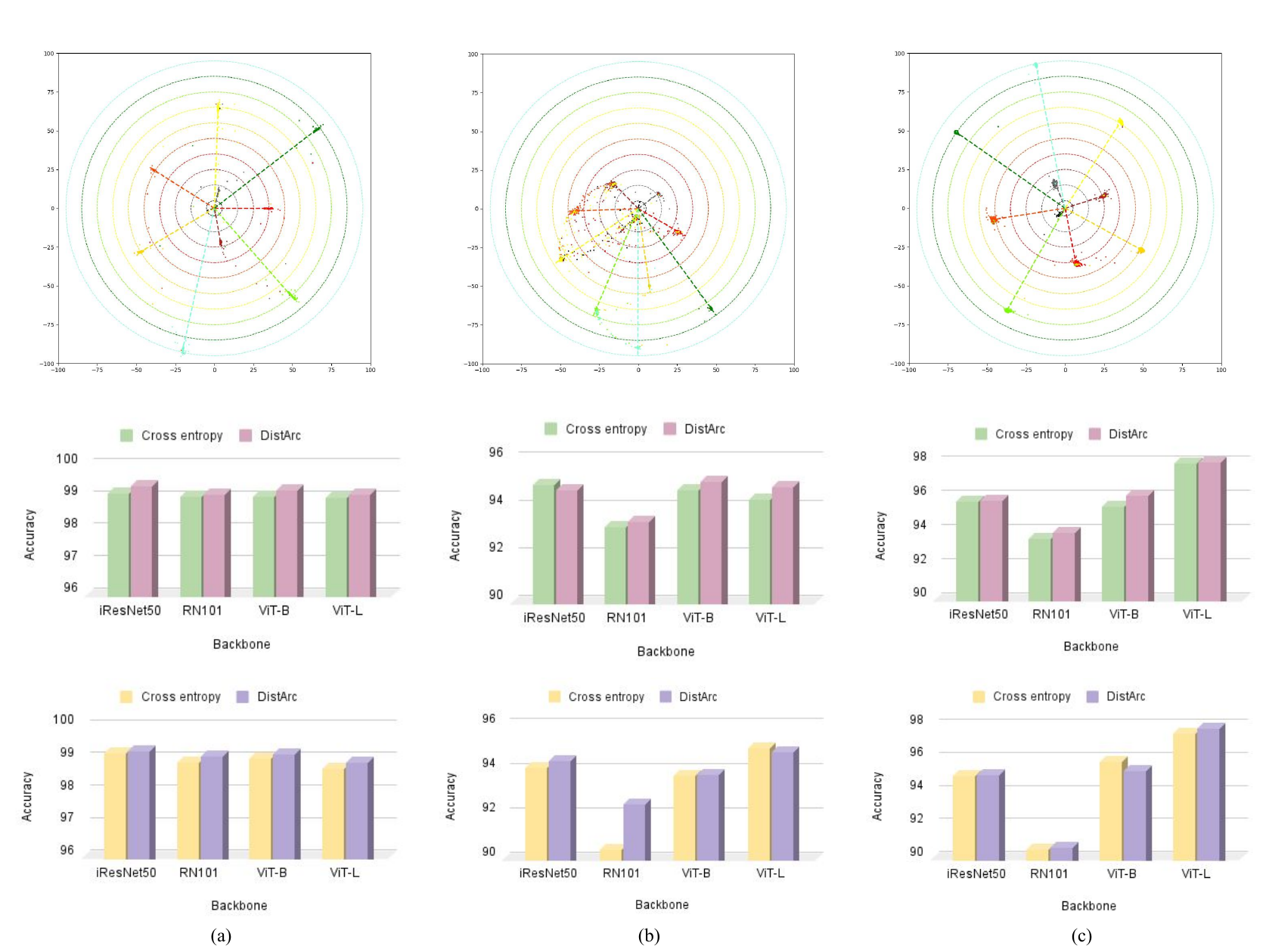}
     \caption{Analysis of loss functions using small-class simple datasets (a) MNIST and (b) FashionMNIST, and complex dataset (c) CIFAR-10. The first row visualizes the features, showcasing the outcomes of learning with \textit{DistArc} loss over 2D multi-spherical manifolds. The last two rows illustrate classification performance of different backbones with a 2-D and 512-D embedding sizes, trained using Cross-entropy and DistArc loss.}
    \label{fig:results_toydata}
\end{figure}

\noindent \textbf{Performance of HyperSpaceX on a Large-Scale Image Dataset:} 
Table \ref{tab:imagenet} provides the performance of the proposed framework after fine-tuning on the ImageNet-1K dataset \cite{deng2009imagenet}. We observe that the proposed DistArc loss outperforms existing methods for embedding sizes of $32$ and $128$. It also achieves the second-highest performance at $512$ dimensions. As the number of classes grows, ArcFace finds it difficult to manage many feature points on a unit hypersphere. This demonstrates the benefit of strategically arranging features in a multi-radial hyperspherical space using DistArc loss during model training.

\begin{table*}[t!]
\centering
\begin{minipage}{.46\textwidth}
\centering
\caption{\label{tab:imagenet}Classification Accuracy (\%) on ImageNet-1K dataset using iResNet50 \cite{duta2021improved}. The best and second best performances are \textbf{bolded} and \underline{underlined}.}
\resizebox{\textwidth}{!}{
\begin{tabular}{|l|c|c|c|}
\hline
\textbf{\begin{tabular}[c]{@{}c@{}}Embedding \\ Size\end{tabular}}          & \textbf{\begin{tabular}[c]{@{}c@{}}Cross\\Entropy\end{tabular}} & \textbf{ArcFace} & \textbf{DistArc} \\ \hline
$\textbf{32}$                        & \underline{48.25}          & 43.05                 & \textbf{55.91}                 \\ \hline
$\textbf{128}$           & \underline{65.18}          & 58.62                & \textbf{65.30}                 \\ \hline
$\textbf{512}$              & \textbf{69.27}          & 59.67                 & \underline{67.90}                 \\ \hline
\end{tabular}}
\end{minipage}
\begin{minipage}{.46\textwidth}
\centering
\caption{\label{tab:ablation_loss_components}Ablation study of \textit{DistArc} loss, reporting classification performance (in $\%$) for various image classification datasets.}
\resizebox{\textwidth}{!}{
\begin{tabular}{|l|c|c|c|}
\hline
\textbf{\begin{tabular}[c]{@{}c@{}}Loss \\ Components\end{tabular}}          & \textbf{MNIST} & \textbf{\begin{tabular}[c]{@{}c@{}}Fashion-\\MNIST\end{tabular}} & \textbf{CIFAR-10} \\ \hline
$\cos(\theta)$                        & 98.97          & 93.69                 & 94.81                 \\ \hline
$\cos(\theta) \; \& \; \cos(\phi)$           & 98.93          & 93.72                 & 95.16                 \\ \hline
$\cos(\theta) \; \& \; \delta$              & 99.01          & 94.83                 & 95.31                 \\ \hline
$\cos(\theta) \; \& \; \cos(\phi) \; \& \; \delta$ & 99.19          & 95.04                 & 96.03                 \\ \hline
\end{tabular}}
\end{minipage}
\end{table*}

\noindent \textbf{HyperSpaceX Performance on Datasets with Fewer Classes:} \cref{fig:results_toydata} showcases the performance and feature visualization for MNIST, FashionMNIST, and CIFAR-10 datasets. The top row of \cref{fig:results_toydata} illustrates the 2D-latent space feature organization across multi-radial hyperspherical manifolds, highlighting the strategic angular positioning. This arrangement emphasizes the significant distance between features of unrelated classes and the closeness of features within the same class, facilitating accurate class predictions. The subsequent rows display a comparison of classification accuracy across different models trained using cross-entropy and DistArc loss functions in $2$D and $512$D latent spaces. These findings illustrate the superior performance of the \textit{HyperSpaceX} framework across both simple (MNIST, FashionMNIST) and more complex (CIFAR-10) datasets with a smaller number of classes.

\noindent \textbf{Training Convergence Analysis:} The training convergence rates of the DistArc loss were benchmarked against existing loss functions. Training with softmax cross-entropy loss shows only a slight decrease in loss values. In contrast, DistArc loss significantly restructures the feature space layout by dispersing identities across various radial hyperspheres. This approach leads to higher initial loss figures that efficiently reduce to a minimum of $0.22$, surpassing the reduction achieved with softmax cross-entropy ($0.34$), CosFace ($2.31$), ArcFace ($1.51$), and OPL ($0.68$) loss functions. A detailed comparison illustrating the training convergence rates of loss functions is available in the supplementary file.

\noindent \textbf{Ablatic Analysis of \textit{DistArc}:} Table \ref{tab:ablation_loss_components} shows the ablation analysis on three image classification datasets, MNIST, FashionMNIST, and CIFAR-10, while utilizing ViT-B architecture (embedding size $512$). The first and second row of Table \ref{tab:ablation_loss_components} shows the efficacy of performance improvement over angular separation through $\theta$ and $\phi$, and the third row shows the improvement with radial and angular factors due to $\delta$ and $\theta$ terms. While, the last row shows the highest performance achieved by including all $\theta$, $\phi$ and $\delta$ components into the loss formulation, highlighting the importance of each component.

\begin{table*}[t!]\renewcommand{\arraystretch}{1.2}
\centering
\caption{\label{tab:results_faces_512} Quantitative results (in \%) of face recognition models trained using different loss functions. The best and second best performances are \textbf{bolded} and \underline{underlined}.}  
\resizebox{0.8\textwidth}{!}{
\begin{tabular}{|l|c|c|c|c|c|}
\hline
\textbf{Loss}    & \textbf{\;\; LFW \;\;}   & \textbf{CFP-FP} & \textbf{AgeDB-30} & \textbf{CA-LFW} & \textbf{CP-LFW} \\ \hline
Triplet          & 98.98          & 91.90           & 89.98             & -               & -               \\
Center loss \cite{centerloss}     & 99.28          & -               & -                 & 85.48      & 77.48      \\ \hline
Softmax          & 99.08          & 94.39           & 92.33             & 88.17       & 84.85       \\
Norm-Softmax     & 98.56          & 89.79           & 88.72             & -               & -               \\ \hline
SphereFace \cite{liu2017sphereface}      & 99.11          & 94.38           & 91.70             & 92.55       & 90.90       \\
CosFace (m=0.35) \cite{wang2018cosface} & 99.10          & \underline{95.44}               & 92.98         & 92.83       & 91.03       \\
ArcFace (m=0.45) \cite{deng2019arcface} & 99.46          & \textbf{95.47}  & \textbf{94.93}    & 92.47       & 90.85       \\
SphereFace2 \cite{Wen2022SphereFace2}     & \underline{99.50}          & -               & 93.68             & \textbf{93.47}           & \underline{91.07}           \\ \hline
DistArc (m=0.4)  & \textbf{99.54} & 95.41           & \underline{94.03}             & \underline{93.32}               & \textbf{91.10}              \\ \hline
\end{tabular}}
\end{table*}

\noindent\textbf{Face Recognition Results:} Table \ref{tab:results_faces_512} showcases detailed comparative results of the \textit{DistArc} loss against other loss functions in face recognition tasks. Utilizing the iResNet50 architecture and training on the CASIA-WebFace dataset, we have evaluated verification performance with a 512-dimensional embedding. The \textit{DistArc} loss demonstrates state-of-the-art (SoTA) results on LFW and CP-LFW datasets and near state-of-the-art results on other datasets.  Additionally, experiments with the iResNet50 backbone on the MS1Mv2 dataset—a derivative of the now-withdrawn MS-Celeb dataset—were conducted to ensure a fair comparison with SoTA methods are included in supplementary materials. The \textit{DistArc} loss trained model achieved following results: 99.82\% on LFW and 98.21\% on AgeDB-30, surpassing results from SphereFace, CosFace, ArcFace, and SphereFace2\footnote{SphereFace (LFW: 99.42\%), CosFace (LFW: 99.73\%), ArcFace (LFW: 99.82\%, AgeDB30: 98.15\%) and SphereFace2 (LFW: 99.50\%, AgeDB30: 93.68\%).}. These findings highlight the \textit{HyperSpaceX} framework's proficiency in crafting highly discriminative and distinct features within the latent space. 

The HyperSpaceX framework is also evaluated on the D-LORD \cite{manchanda2023d} dataset, a large-scale open-set surveillance dataset with $600$ test subjects, using a deep metric learning approach \cite{oh2016deep}. With ArcFace, a Rank-1 identification accuracy of $61.84\%$ was achieved, while DistArc achieved $62.71\%$. For Rank-5 accuracy, the results were $65.89\%$ and $66.03\%$, respectively, indicating that DistArc is more effective at learning enhanced discriminative features than ArcFace.

\section{Conclusion}
\label{sec:conclusion}
In this research, we introduce the \textit{HyperSpaceX} framework, which navigates both angular and radial dimensions, facilitating a uniquely distinguishable and discernible arrangement of class features within multi-hyperspherical manifold space. This is achieved using the novel \textit{DistArc} loss, which leverages the radial and angular components of the feature space. This loss function significantly enhances the cohesion among similar class features while concurrently maximizing the distances between different classes, resulting in more precise and discriminative feature representation. To evaluate the performance of this radial-angular framework, we introduce a predictive measure that utilizes the shortest resultant vector between the embeddings of test samples and proxy vectors, offering a deeper insight into the model's performance. The efficacy of \textit{HyperSpaceX} is demonstrated through comprehensive experiments on seven image classification datasets and six face recognition datasets. The proposed methodology shows improvements in handling a diverse array of image and face datasets, ranging from simpler to more complex and large-scale collections encompassing several classes and samples.

\section*{Acknowledgements}
Chiranjeev and Thakral received partial support from the PMRF Fellowship, and Vatsa is partially supported by the Swarnajayanti Fellowship.


%
%

\bibliographystyle{splncs04}
\bibliography{main}
 
\end{document}


\let\titleold\title
\renewcommand{\title}[1]{\titleold{#1}\newcommand{\thetitle}{#1}}
\title{HyperSpaceX: Radial and Angular Exploration of HyperSpherical Dimensions \\  Supplementary Material} 
\titlerunning{HyperSpaceX}
\author{Chiranjeev Chiranjeev\orcidlink{0000-0003-3026-1255} \and
Muskan Dosi\orcidlink{0000-0001-7451-3317} \and
Kartik Thakral\orcidlink{0000-0002-2528-9950} 
Mayank Vatsa\orcidlink{0000-0001-5952-2274} \and
Richa Singh\orcidlink{0000-0003-4060-4573}}

\authorrunning{Chiranjeev et al.}

\institute{Indian Institute of Technology Jodhpur, India
\email{\{chiranjeev.1,dosi.1,thakral.1,mvatsa,richa\}@iitj.ac.in} \\
\textcolor{magenta}{\url{https://github.com/IAB-IITJ/HyperSpaceX}}}


\maketitle

\setcounter{page}{1}
\setcounter{section}{0}
\setcounter{figure}{0}
\setcounter{table}{0}
\setcounter{equation}{0}

\section{Derivation of Decision Boundary and Predictive Measure for HyperSpaceX}
\label{sec:derivation}
The \textit{HyperSpaceX} framework depends on three primary components: $\theta$, $\phi$, and $\delta$ for introducing more refined decision boundaries for separability among classes and more discriminative representation learning of features. We derive the formulations dependent on these three essential components of the HyperSpaceX framework for two purposes: $(1)$ for the decision boundary and $(2)$ predictive measure for the most favourable class distribution. The formulations are built using cosine angular-based triangle law properties. In which triangular formation is created between three vectors: scaled proxy vector $\omega_{r}$, sample's feature vector $x$, and resultant vector $R$ between $\omega_{r}$ and $x$, having angle $\theta$ between $\omega_{r}$ and $x$, and angle $\phi$ between $\omega_{r}$ and $R$. 

\noindent\textbf{(1) Decision Boundary:} Assuming a scenario of $2$-class classification problem. The decision boundary dependent on three essential loss factors is derived as follows:

\begin{equation}
\label{eq:decision_eq}
    \begin{aligned}
    (\cos(\theta_1 + m) + ||x_1||_{2}) - (\cos(\theta_2) + ||x_2||_{2}) &= 0 \text{ for class 1} \notag \\
    (\cos(\theta_1) + ||x_1||_{2}) - (\cos(\theta_2 + m) + ||x_2||_{2}) &= 0 \text{ for class 2}
    \end{aligned}
\end{equation}

\noindent where, according to cosine angular-based triangle law $||x_1||_{2}$ is defined as,

\begin{equation}
\label{eq:decision_eq1}
    \begin{aligned}
    ||R_1||_{2}\cos(\theta_1 + m) + ||\omega_{r_1}||_{2}\cos(\pi - ((\theta_1 + m)+\phi_1))
    \end{aligned}
\end{equation}
and $||x_2||_{2}$ as, 
\begin{equation}
\label{eq:decision_eq2}
    \begin{aligned}
    ||R_2||_{2}\cos(\theta_2 + m) + ||\omega_{r_2}||_{2}\cos(\pi - ((\theta_2 + m)+\phi_2))
    \end{aligned}    
\end{equation}

\noindent\textbf{(2) Predictive Measure:} The class prediction is made based on the smallest resultant vector being selected from a set of resultant vectors computed between scaled proxy and test sample embedding. The resultant vector $(R_{i})$ and its magnitude computation for a test sample feature $(x)$ is calculated against each scaled proxy vector $(\omega_{r_{i}})$ using the vector formulation,
\begin{equation}
\label{eq:evaluation_R}
    \begin{aligned}
        ||R_{i}||_{2} = || x - \omega_{r_{i}} ||_{2}  \;\;\;\; i \in 1,2,3,... 
    \end{aligned}
\end{equation}
and, resultant vector's magnitude calculation using cosine angular-based triangle side length computation property, showing its dependence on the introduced angular and radial components of HyperspaceX framework $\ie$, $\theta$, $\phi$ and length of test sample feature vector $||x||_{2}$,

\begin{equation}
\label{eq:evaluation_R_magnitude}
    \begin{aligned}
    ||R_{i}||_{2} = ||x||_{2} \, \cos\phi_{i} \;+\; ||\omega_{r_{i}}||_{2} \,
    \cos(\pi-(\theta_{i}+\phi_{i}))
    \end{aligned}
\end{equation}

where, $(\pi-(\theta_{i}+\phi_{i})$, represents the value of the angle between $x$ and $R$, which is calculated from all interior angles in a triangle sums to $180$\textdegree or $\pi$ in radians.

\section{Geometric Visual and Experimental Ablation of DistArc Loss in HyperSpaceX Methodology}
\label{sec:visual_experimental_ablation}

\begin{figure}[t]
\centering
\begin{subfigure}{.49\linewidth}
 \centering
    \includegraphics[width=\linewidth]{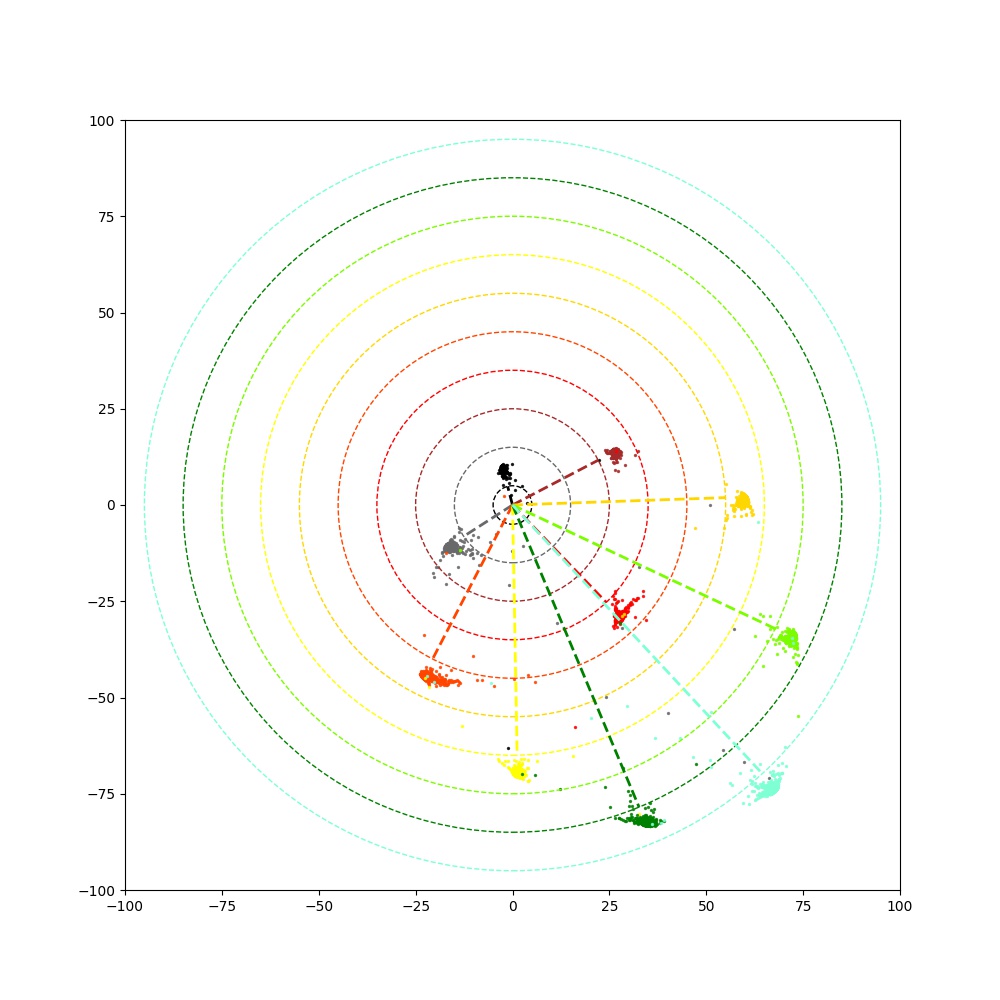}
    \caption{}
    \label{fig:costheta_cosphi}
\end{subfigure}
\begin{subfigure}{.49\linewidth}
\centering
    \includegraphics[width=\linewidth]{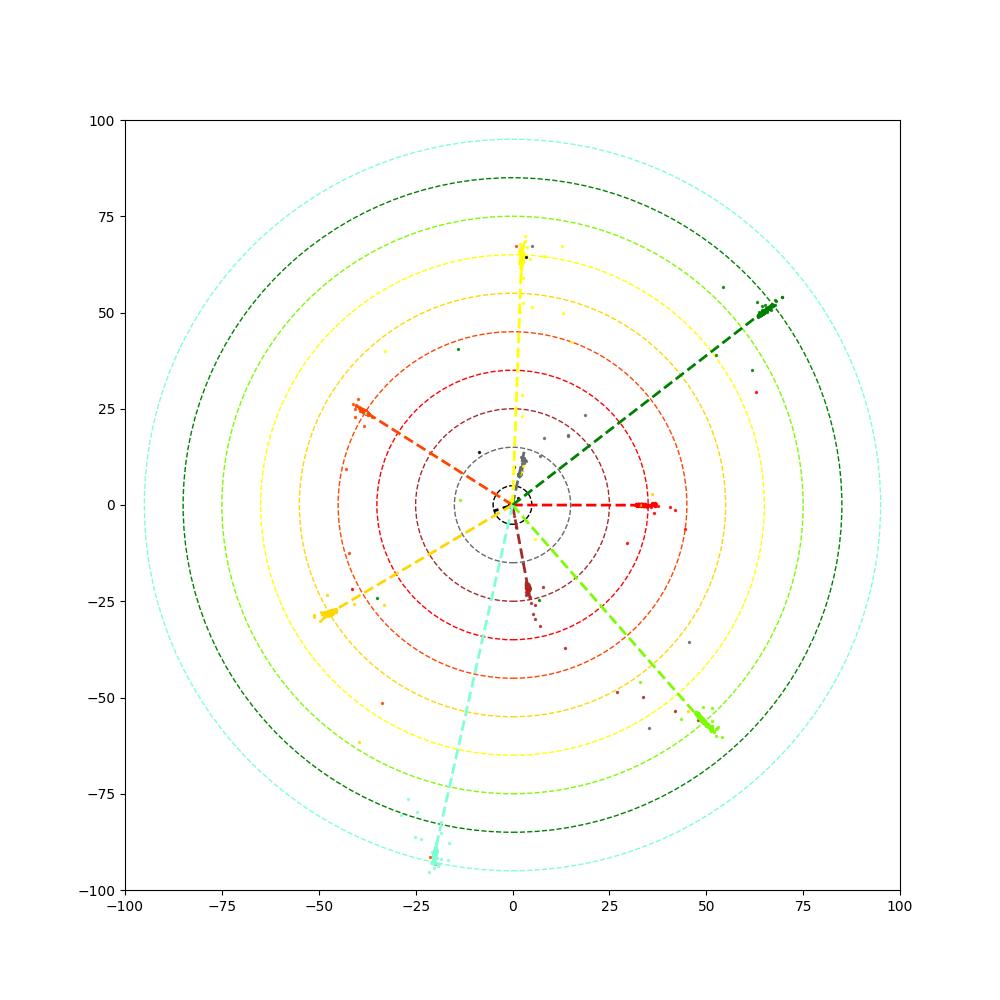}
    \caption{}
    \label{fig:dist_costheta_cosphi}
\end{subfigure}

\caption{Visualization of geometric ablation of feature representation learning via DistArc loss components. This image shows the significance of (a) $\cos\theta$ and $\delta$, and (b) the incorporation of $\cos\phi$ in DistArc loss formulation. Part (a) illustrates the distribution and grouping of features in both angular and radial dimensions. Part (b) further refines this concept by displaying enhanced feature compactness and the scaling of features to specific radii-hyperspheres, each associated with their class proxies in radial-angular dimensions. The diverse color spectrum represents unique class distributions, each aligned with their scaled class proxies, depicted as colored lines in distinct angular orientations. The overlay of concentric circles symbolizes the 2D-hyperspherical manifolds, each with varying radii, underlining the multi-dimensional aspect of this learning model.}
\label{fig:gemoetric_feature_visual}
\end{figure}

This section provides a geometric and experimental interpretation of each component of the proposed \textit{DistArc} loss by learning in both combined angular and radial dimensions. Fig. \ref{fig:gemoetric_feature_visual} represents the overall geometric visualization of all parts of the proposed loss function $\ie$, $\cos\theta$, $\cos\phi$ and $\delta$.  

\noindent\textbf{Visual Geometric Ablation:} The geometric interpretation of both components, $\cos(\theta_{y_{i}}+m)$,  $\delta_{y_{i}}$ and $\delta_{j}$ in the proposed \textit{DistArc} loss function is visually represented through the depiction of features in a 2-dimensional latent hyperspherical space, as illustrated in Fig. \ref{fig:costheta_cosphi}. This representation demonstrates that these terms help position each class feature to the particular radial hypersphere along with the minimization of angles $\theta$. These $\delta$-based loss components help increase separability among different classes in radial dimensions, and discriminability is also achieved by further minimizing intra-class distance among similar class features and maximizing inter-class distance among distinct classes.  

Fig. \ref{fig:dist_costheta_cosphi} shows the significance of all components along with $\cos\phi_{y_{i}}$ term in the DistArc loss. Feature representation in Fig. \ref{fig:dist_costheta_cosphi} interprets that these terms help create a more compact distribution of each class cluster along with keeping the dissimilar class features apart by distance/radial factor. We can observe that the minimization of angles $\theta$ and $\phi$ leads to a concentrated angular distribution of features and clustering among features belonging to the same class. Furthermore, it prevents features from extending beyond their respective hypersphere of radii $||\omega_{r_{y_{i}}}||_{2}$. The angular separability and discriminability among classes are achieved through these cosine-based terms in the loss.

The \textit{HyperSpaceX} framework, as observed through the geometric visualization of feature representation learned using DistArc loss in Fig. \ref{fig:gemoetric_feature_visual}, leads to the enlargement and alignment of feature vectors to their respective angular and radial positions within the hyperspheres space. This results in a more separable and discriminative arrangement of data points across various hyperspherical dimensions. The overall combination of angular loss minimization due to $\theta$ and $\phi$, and angular-radial loss minimization due to $\delta$ and $\phi$ helps to increase the intra-class similarity among same class features at point $\omega_{r_{y_{i}}}$ by shifting the features up to the $||\omega_{r_{y_{i}}}||_{2}$ radial hypersphere and compacting it closer to $\omega_{r_{y_{i}}}$ scaled proxy vector. Therefore, the DistArc loss ensures that features $x_{i}$ maintain specific angular orientations and prevent them from extending beyond their respective radial hyperspheres. Additionally, it elevates inter-class distance among features $x_{i}$ and the dissimilar class proxies $\omega_{j}$\textit{'s}. This approach, incorporated within the HyperSpaceX methodology, contributes to enhancing feature discriminability and separability.

\begin{figure}[t]
\centering
\begin{subfigure}{.49\linewidth}
 \centering
    \includegraphics[width=\linewidth]{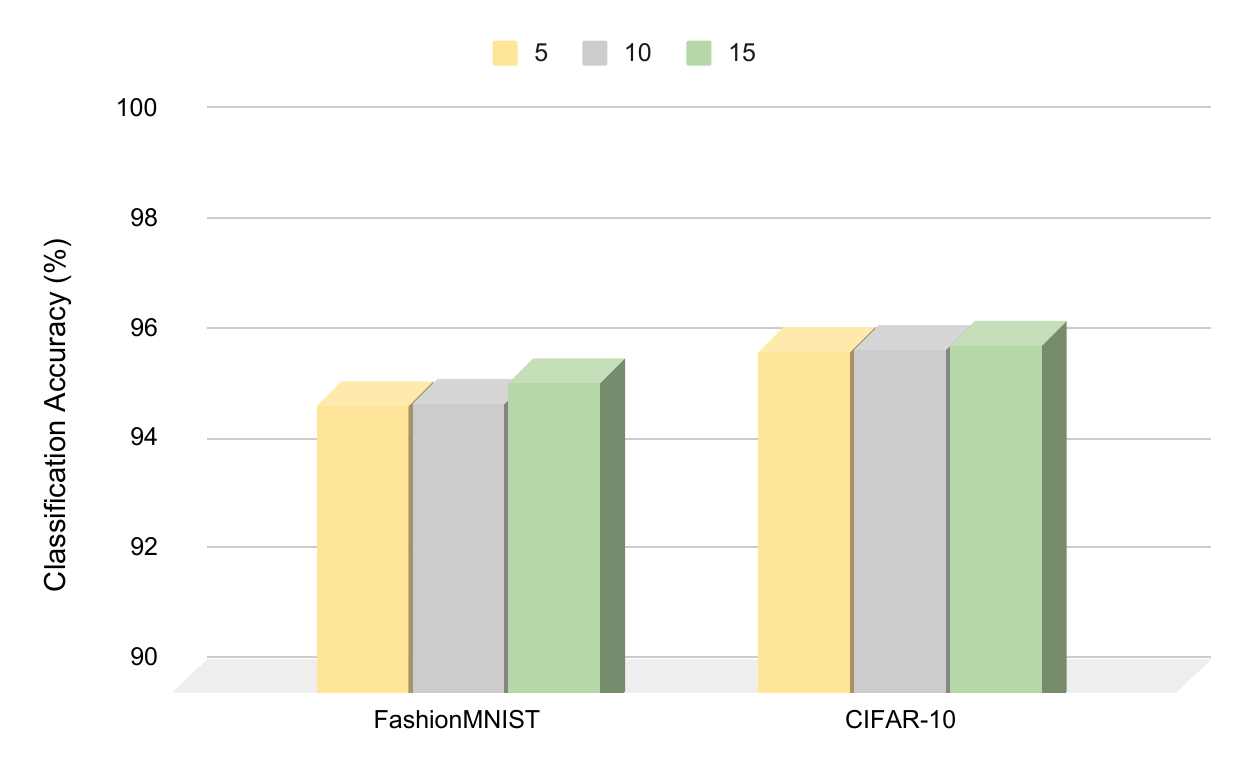}
    \caption{}
    \label{fig:ablation_gap}
\end{subfigure}
\begin{subfigure}{.49\linewidth}
\centering
    \includegraphics[width=\linewidth]{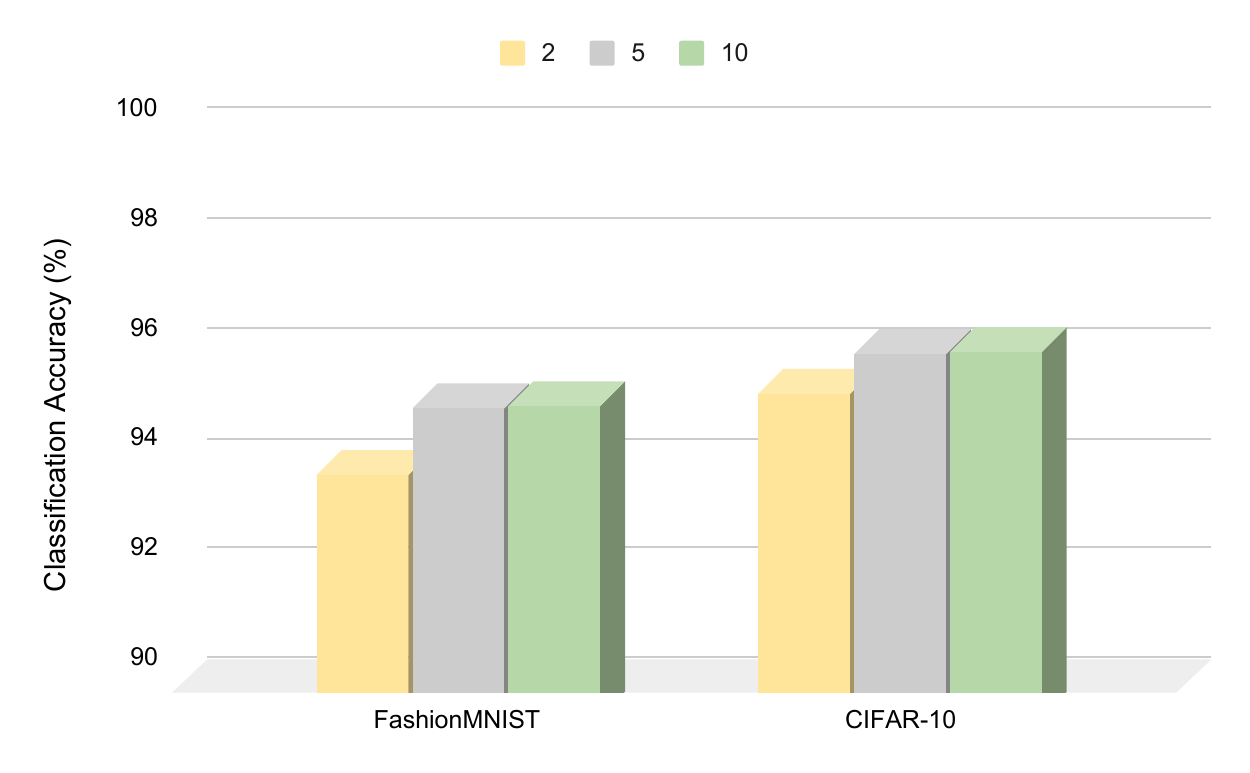}
    \caption{}
    \label{fig:ablation_number_hyperspheres}
\end{subfigure}

\caption{Classification performance (in $\%$) (a) by varying radial gap between adjacent hyperspheres where each color denotes gap between the hyperspheres; (b) by varying number of hyperspheres where each color denotes the different number of hyperspheres chosen.}
\label{fig:ablation_study_experimental_results}
\end{figure}

\noindent\textbf{Experimental Ablation:} Experimentally, we highlight the importance of each field introduced in the proposed \textit{HyperSpaceX} framework. We reported extensive results showcasing the classification accuracy on various image classification datasets while utilizing ViT-B architecture with an embedding layer of size $512$ dimensions. From Fig. \ref{fig:ablation_gap}, we can analyze the performance increment as we increase the radial gap between adjacent hyperspheres. It illustrates as the gap increases, the separability among the different classes increases. It leads to reduced miss-classifications among class feature points of distinct classes in the latent space. Fig. \ref{fig:ablation_number_hyperspheres} represents the classification performance on varying the number of hyperspheres, for distributing the class features over different hyperspherical regions. It can be analyzed as we increase the number of hyperspheres; it leads to more enhanced separability and discriminability among class features. This causes performance improvement as we increase the angular and radial search space exploration through multi-hyperspherical regions. 

Consequently, from both Visual and Experimental Ablation study, we can highlight the key significance of \textit{HyperSpaceX} framework for learning highly separable and discriminative representation of features in latent space using \textit{DistArc} loss. In the proposed DistArc loss function, each introduced component plays a crucial role in enhancing performance across multiple datasets by representing feature points in various angular and radial dimensions. All components are carefully designed to optimize the function's effectiveness, demonstrating their individual significance in diverse data environments.
\begin{figure}[!t]
    \centering
    \includegraphics[width=0.7\textwidth]{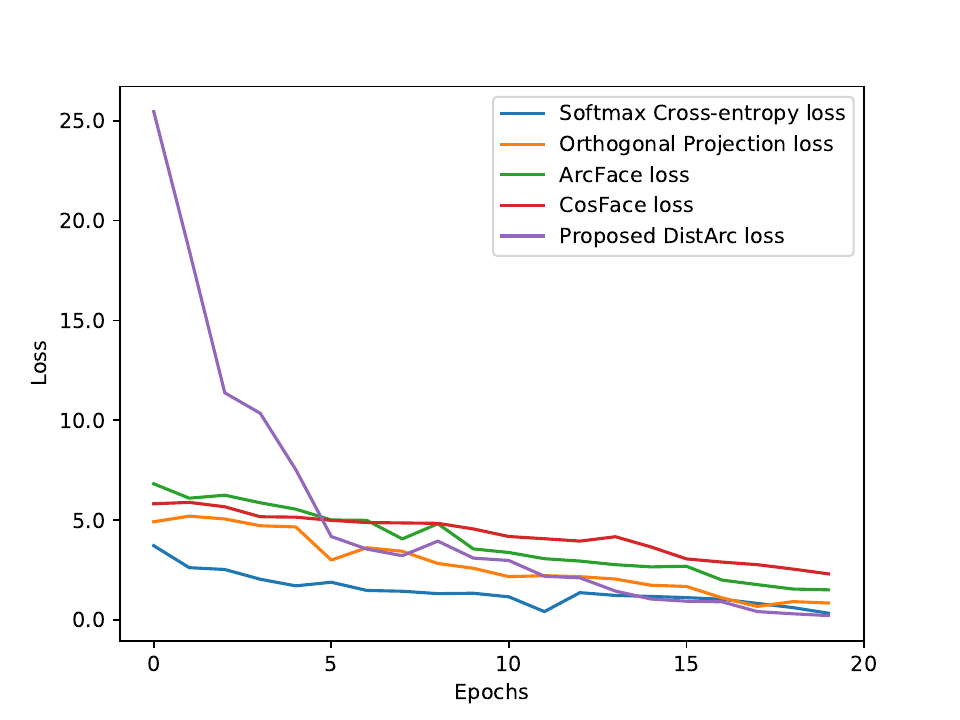}
    \caption{Comparative training convergence graph across different loss formulations.}
    \label{fig:train_complexity}
\end{figure}

\section{Graphical Representation of Training Convergence}
The results of various loss formulations and training convergence are depicted through the graph in Fig. \ref{fig:train_complexity}. It shows the loss convergence over epochs. It can be visualized that the DistArc loss leads to higher loss values in initial epochs, which decrease gradually due to the diverging of each identity to different radial hyperspheres. It can be observed that the DistArc loss curve converges faster in comparison to ArcFace, CosFace, and Orthogonal Projection losses, which further finally converges to the lowest minimum value of $0.22$ in comparison to other loss functions even including cross-entropy loss, which attains a minimum of $0.34$ loss value. 

\begin{table*}[t!]\renewcommand{\arraystretch}{1.2}
\centering
\caption{\label{tab:ms1mv2_suppl_results} Quantitative results (in \%) of face recognition model trained using different loss functions on MS1Mv2 dataset. The best and second best performances are \textbf{bolded} and \underline{underlined}.} 
\resizebox{0.7\textwidth}{!}{
\begin{tabular}{|l|c|c|c|c|}
\hline
\textbf{Loss}    & \textbf{ \;       LFW     \;    }    & \textbf{AgeDB-30} & \textbf{CA-LFW} & \textbf{CP-LFW} \\ \hline
Center loss       & 98.75                        & -                 & 85.48               & 77.48            \\ \hline
SphereFace       & 99.42                        & 92.88                 & 90.30               & 81.40               \\
CosFace & 99.73                      & 92.98                 & 92.83               & 91.03               \\
ArcFace  & 99.82               & \underline{98.15}             & \textbf{95.45}               & \textbf{92.08}               \\
SphereFace2      & \underline{99.50}                       & 93.68             & 93.47               & 91.07               \\ \hline
DistArc  & \textbf{99.82}               & \textbf{98.21}    & \underline{94.87}               & \underline{91.10}              \\ \hline
\end{tabular}}
\end{table*}

\section{More Empirical Results}
\noindent\textbf{Face Recognition:} We conducted another experiment by training the model on one of the largest face datasets, MS1Mv2. This experiment is conducted to ensure a fair comparison with state-of-the-art (SoTA) methods that have also been trained on this dataset. Table \ref{tab:ms1mv2_suppl_results} presents the performances of several face recognition techniques on testing face databases. It can be observed that the model when trained with the proposed DistArc loss on the MS1Mv2 dataset, achieves superior performance on most of the testing face datasets. This demonstrates the efficacy of the proposed DistArc Loss in large-scale face datasets as well. 

\begin{table*}[t!]\renewcommand{\arraystretch}{1.1}
\centering
\caption{\label{tab:larger1024dim}Image classification accuracy (in \%) of models trained on $1024$ dimensions with various loss functions. The best and second best performances are \textbf{bolded} and \underline{underlined}.} 
\resizebox{0.7\linewidth}{!}{
\begin{tabular}{lcccc}
\hline
\multicolumn{5}{|c|}{\textbf{CIFAR-100}} \\ \hline
\multicolumn{1}{|l|}{\textbf{Backbone}} & \multicolumn{1}{c|}{\textbf{Cross entropy}} & \multicolumn{1}{c|}{\;\;\;\;\textbf{ArcFace}                    \;\;\;\;\;} & \multicolumn{1}{c|}{\;\;\;\;\textbf{CosFace}     \;\;\;\;\;} & \multicolumn{1}{c|}{\;\;\;\; \textbf{DistArc} \;\;\;\;\;} \\ \hline
\multicolumn{1}{|l|}{\textbf{iResNet50}} & \multicolumn{1}{c|}{\underline{85.16}} & \multicolumn{1}{c|}{82.67} & \multicolumn{1}{c|}{83.11} & \multicolumn{1}{c|}{\textbf{89.52}} \\
\multicolumn{1}{|l|}{\textbf{ViT-B}} & \multicolumn{1}{c|}{\underline{82.41}} & \multicolumn{1}{c|}{81.00} & \multicolumn{1}{c|}{80.84} & \multicolumn{1}{c|}{\textbf{85.74}} \\ \hline 
\multicolumn{5}{|c|}{\textbf{TinyImageNet}} \\ \hline

\multicolumn{1}{|l|}{\textbf{Backbone}} & \multicolumn{1}{c|}{\textbf{Cross entropy}} & \multicolumn{1}{c|}{\textbf{ArcFace}} & \multicolumn{1}{c|}{\textbf{CosFace}} & \multicolumn{1}{c|}{\textbf{DistArc}} \\ \hline
 \multicolumn{1}{|l|}{\textbf{iResNet50}} & \multicolumn{1}{c|}{60.24} & \multicolumn{1}{c|}{61.47} & \multicolumn{1}{c|}{\underline{62.01}} & \multicolumn{1}{c|}{\textbf{64.01}} \\
 \multicolumn{1}{|l|}{\textbf{ViT-B}} & \multicolumn{1}{c|}{\textbf{78.26}} & \multicolumn{1}{c|}{77.96} & \multicolumn{1}{c|}{78.07} & \multicolumn{1}{c|}{\underline{78.22}} \\ \hline
\end{tabular}}
\end{table*}

\noindent\textbf{Image Classification:} The classification results with $1024$ embedding size on CIFAR-100 and TinyImageNet datasets are shown in \textbf{Table \ref{tab:larger1024dim}}. We can observe superior performance on backbones trained with DistArc loss. Further, an improvement of up to $4.36\%$ on CIFAR-100 and about $2.0\%$ on TinyImageNet datasets for models trained with the proposed DistArc loss and other existing loss functions. We have also conducted \textbf{McNemar's test} for iResNet50 backbone on TinyImageNet dataset with $512$ embedding size, which yields $\chi^{2} = 4.848$ ($>$$\chi^{2}_{0.05}$), stating a significant difference between models trained using cross-entropy loss and proposed DistArc loss. 

\begin{figure}[t!]
    \centering
    \includegraphics[width=\textwidth]{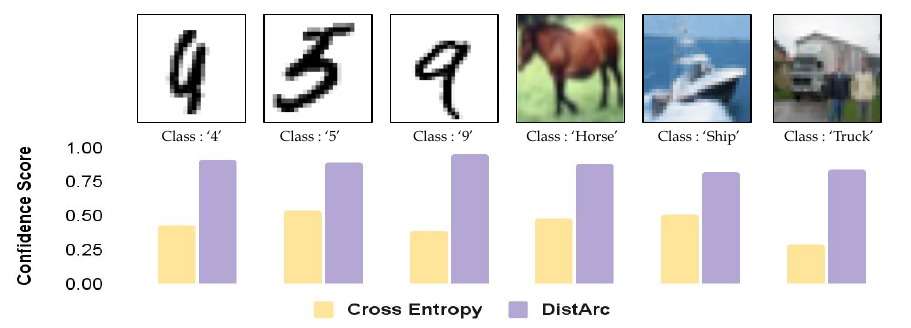}
    \caption{Analytical results demonstrating the comparative study of confidence scores in correct class predictions. The first row shows the images from two datasets, MNIST and CIFAR-10, along with their ground truth class. The second row illustrates the corresponding confidence scores of class predictions made by models trained with Cross entropy loss and DistArc loss.}
    \label{fig:confidence_scores_fig}
\end{figure}

\noindent\textbf{Confidence in Class Predictions:} The quality of predictions should be evaluated not only based on their correctness but also on the level of confidence associated with them. In Fig. \ref{fig:confidence_scores_fig}, we compare the confidence levels of predictions made using a model trained with cross-entropy loss and DistArc loss. For each sample depicted in Fig. \ref{fig:confidence_scores_fig}, we consistently observe that predictions made using DistArc loss exhibit significantly higher confidence for both the MNIST and CIFAR-10 datasets. These experiments show that feature points near the decision boundary have lower prediction confidence with cross-entropy loss. With the proposed DistArc loss, features are compact within the same class and separative between distinct classes, leading to high predictive scores. This demonstrates the effectiveness of introducing the HyperSpaceX framework that primarily focused on separating the distinct classes with significantly larger distances and compacting the same class features, which is achieved through the proposed DistArc loss, hence resulting in producing confident classifications. We also applied a temperature-based calibration \cite{guo2017calibration} technique for better confidence score understanding. After calibrating the iResNet50 backbone trained on the CIFAR-100 dataset using Cross-entropy and DistArc loss, we achieve an Expected Calibration Error (ECE) of $0.467$ and \pmb{$0.337$}, respectively (lower is better).

\begin{table}[!t]
\centering
\caption{\label{tab:radii} Classification results across different radial factors $(r_{y_{i}})$ on CIFAR-10 dataset.}
\resizebox{0.75\linewidth}{!}{
\begin{tabular}{|c|c|c|c|}
\hline
\textbf{radius $(r_{y_{i}})$}           & \textbf{1,2,3,…,10} & \textbf{5,10,15,…,50} & \textbf{10,20,30,…,100} \\ \hline
\textbf{Accruacy (in \%)} & 94.18                 & 95.48                     & 95.82                       \\ \hline
\end{tabular}}
\end{table}

\noindent\textbf{Sensitivity analysis based on $r_{y_{i}}$:} Since $r_{y_i}$ refers to a scaling radial factor defining the radials of multi-hyperspheres. We have performed its sensitivity analysis on the CIFAR-10 dataset with iResNet50 backbone through Table \ref{tab:radii}. It can be analyzed that larger radial values and gaps lead to more multi-spherical feature space exploration. It results in more separated distinct class features in the latent space, causing reduced inter-mixing of feature points. Hence, it leads to more enhanced discriminative features, causing improvement in classification performance as the radial gap increases.

\begin{table}[!t]
\centering
\caption{\label{tab:lambda_exp} Ablative experiment for $\lambda$ on CIFAR-10 dataset with iResNet50 backbone.}
\resizebox{0.5\linewidth}{!}{
\begin{tabular}{|c|c|c|c|c|}
\hline
\pmb{$\lambda$} & \textbf{0.001} & \textbf{0.002} & \textbf{0.005} & \textbf{0.01} \\ \hline
\textbf{Accruacy (in \%)}       & 94.55              & 95.17              & 95.82             & 94.07             \\ \hline
\end{tabular}}
\end{table}

\noindent\textbf{Effect of $\lambda$:} Through Table \ref{tab:lambda_exp}, we study the effect of $\lambda$ as we vary the values of $\lambda$ by training on CIFAR-10 dataset. It can be observed that HyperSpaceX performs optimally with $\lambda$ as a multiplier of $1\mathrm{e}{-3}$, achieving a balanced DistArc loss and highly discriminative feature learning.

\begin{table*}[t!]\renewcommand{\arraystretch}{1.2}
\centering
\caption{\label{tab:image_data_stats}Statistics of the used datasets for the Image classification task.} 
\resizebox{0.9\textwidth}{!}{
\begin{tabular}{|l|c|c|c|}
\hline
\textbf{Dataset} & \textbf{Total \# of images} & \textbf{\# of training images} & \textbf{\# of test images} \\ \hline
MNIST            & 60K                         & 50K                            & 10K                        \\
FashionMNIST     & 70K                         & 60K                            & 10K                        \\
CIFAR-10         & 60K                         & 50K                            & 10K                        \\
CIFAR-100        & 60K                         & 50K                            & 10K                        \\
TinyImageNet     & 0.11M                       & 0.1M                           & 10K                        \\
CUB-200          & 11,788                      & 5,994                          & 5,794                      \\
ImageNet         & 1.38M                       & 1.28M                          & 0.1M                      \\ \hline
\end{tabular}}
\end{table*}

\begin{table*}[t!]\renewcommand{\arraystretch}{1.2}
\centering
\caption{\label{tab:face_data_stats}Statistics for the used datasets for Face Recognition task.} 
\resizebox{0.7\textwidth}{!}{
\begin{tabular}{|l|c|c|c|}
\hline
\textbf{Dataset} & \textbf{\# of identities} & \textbf{\# of images} & \textbf{\;\;\;\;\;split\;\;\;\;\;} \\ \hline
CASIA-WebFace    & 10K                       & 0.5M                  & train          \\
MS1Mv2           & 87K                       & 5.8M                  & train           \\
D-LORD           & 1,500                     & 0.22M                 & train           \\ \hline
LFW              & 5,749                     & 13,233                & test           \\
CFP-FP           & 500                       & 7,000                 & test           \\
AgeDB-30         & 568                       & 16,488                & test           \\
CA-LFW           & 5,749                     & 11,652                & test           \\
CP-LFW           & 5,749                     & 12,174                & test           \\ 
D-LORD           & 600                       & 36,420                & test           \\
\hline
\end{tabular}}
\end{table*}

\section{Limitations and Future Work}
We note certain limitations of the HyperSpaceX framework that will be further addressed in future work. A flexible learnable margin can be introduced in the proposed HyperSpaceX framework to further reduce the inter-mixing of feature points of distinct classes over consecutive hyperspheres, \ie enhancing more discriminability among features. Further, as the number of radials increases extremely, our approach necessitates the adoption of a learnable, un-normalized proxy expansion technique to ensure smoother learning and reduced feature spreading while expanding the radials to a greater extent. Meanwhile, our method excels in distinctive class learning and classification enhancement, especially in the visual domain by improving classification and recognition performance, its adoption to other modalities like text and audio currently remains unexplored.

\section{Statistics for the Used Datasets}
Table \ref{tab:image_data_stats} and Table \ref{tab:face_data_stats} represent the dataset statistics of the image classification and face recognition datasets, respectively. The face identification performance on D-LORD dataset is evaluated for high-resolution face images up to a distance of 5m. Overall these datasets are utilized to conduct training and testing experiments for the proposed and existing methodologies. 

\bibliographystyle{splncs04}
\bibliography{main}